\documentclass[a4paper,10pt,oneside,fleqn,review]{cas-dc}

\usepackage[authoryear]{natbib}
\usepackage{array, longtable, tabularx}
\usepackage{pdflscape}
\usepackage{xltabular}
\usepackage{afterpage}
\usepackage{cas-common}
\usepackage{multicol}
\def\tsc#1{\csdef{#1}{\textsc{\lowercase{#1}}\xspace}}
\newcolumntype{C}[1]{>{\centering\arraybackslash}p{#1}}
\tsc{WGM}
\tsc{QE}

\begin{document}
\let\WriteBookmarks\relax
\def\floatpagepagefraction{1}
\def\textpagefraction{.001}

\shorttitle{Sugarcane Health Monitoring With Satellite Spectroscopy and Machine Learning: A Review}    

\shortauthors{Ethan Kane Waters, Carla Chia-ming Chen and Mostafa Rahimi Azghadi}  

\title [mode = title]{Sugarcane Health Monitoring With Satellite Spectroscopy and Machine Learning: A Review}  

\author[1,2]{Ethan Kane Waters}[orcid=0000-0003-0624-4877]
\corref{cor1}
\ead{ethan.waters@jcu.edu.au}
\credit{Conceptualization, Methodology, Investigation, Visualization, Writing – original draft \& editing.}

\affiliation[1]{organization={College of Science and Engineering},
            addressline={James Cook University}, 
            city={Townsville},
            postcode={4818}, 
            state={QLD},
            country={Australia}}

\affiliation[2]{organization={Agriculture Technology and Adoption Centre},
            addressline={James Cook University}, 
            city={Townsville},
            postcode={4814}, 
            state={QLD},
            country={Australia}}

\author[1,2]{Carla Chia-ming Chen}[orcid=0000-0001-9718-4464]
\ead{carla.ewels@jcu.edu.au}
\credit{Supervision, Writing – critical review \& extensive editing}

\author[1,2]{Mostafa Rahimi Azghadi}[orcid=0000-0001-7975-3985]
\corref{cor1}
\ead{mostafa.rahimiazghadi@jcu.edu.au}
\credit{Conceptualization, Supervision, Writing – critical review \& extensive editing}

\cortext[cor1]{Corresponding author}

\begin{abstract}
Research into large-scale crop monitoring has flourished due to increased accessibility to satellite imagery. This review delves into previously unexplored and under-explored areas in sugarcane health monitoring and disease/pest detection using satellite-based spectroscopy and Machine Learning (ML). It discusses key considerations in system development, including relevant satellites, vegetation indices, ML methods, factors influencing sugarcane reflectance, optimal growth conditions, common diseases, and traditional detection methods. Many studies highlight how factors like crop age, soil type, viewing angle, water content, recent weather patterns, and sugarcane variety can impact spectral reflectance, affecting the accuracy of health assessments via spectroscopy. However, these variables have not been fully considered in the literature. In addition, the current literature lacks comprehensive comparisons between ML techniques and vegetation indices. We address these gaps in this review.
We discuss that, while current findings suggest the potential for an ML-driven satellite spectroscopy system for monitoring sugarcane health, further research is essential. This paper offers a comprehensive analysis of previous research to aid in unlocking this potential and advancing the development of an effective sugarcane health monitoring system using satellite technology.

\end{abstract}

\begin{keywords}
Sugarcane\sep Health Monitoring System\sep Remote Sensing\sep Satellites\sep Spectroscopy\sep Machine Learning\sep Vegetation Indices\sep Disease\sep Pests\sep 

\end{keywords}

\maketitle

\begin{figure*}[]
    \centering
        \includegraphics[width=\textwidth]{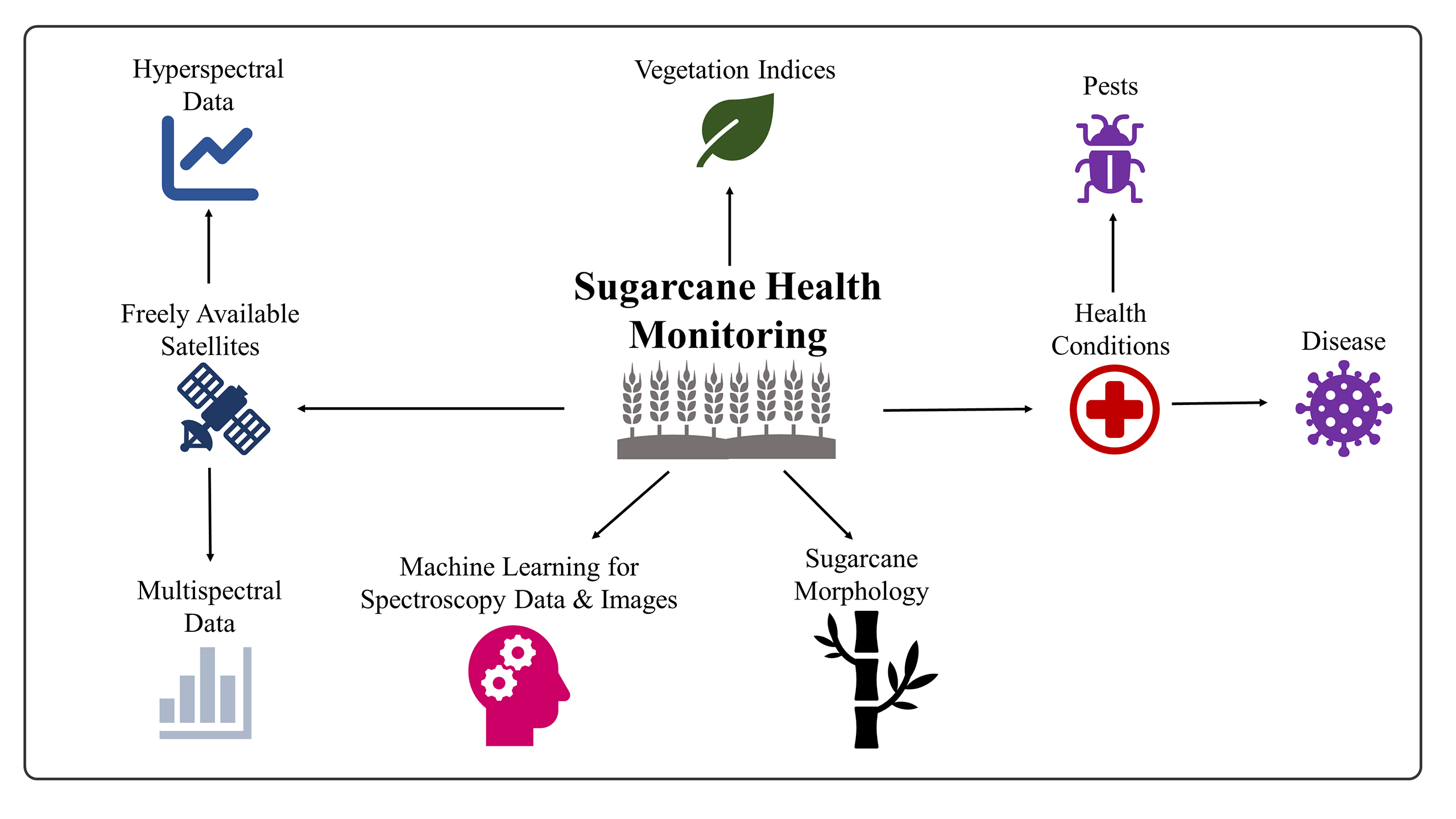}
    \caption{Qualitative relationship between the components necessary to perform sugarcane health monitoring with spectroscopy. \label{fig14}}
\end{figure*}


\section{Introduction}

The Earth’s population has rapidly increased from 2.5 billion in 1950 to over 8 billion in 2024. This change has reflected a significant increase in the aggregate agricultural production to sustain the population \citep{RN1,RN2,RN3}. The global population growth is projected to reach 9.15 billion by 2050 and is expected to require a 70\% increase in aggregate agricultural production from 2012  \citep{RN1, RN2, RN4, RN5, RN6}. It has become increasingly difficult to increase yield and agricultural production with traditional methods and this has led to increased demand for smart agricultural practices \citep{RN1}. 

Sugarcane is one of the world's most important agricultural products, which is grown and produced in large volumes. For instance, it was reported that in 2021, global sugarcane production was 1861.9 million metric tons. Australia produced 31.13 million metric tonnes in 2021 \citep{RN7}, putting it ninth for worldwide production, where 95\% of production occurred in the state of Queensland. Sugarcane yield can be heavily hampered by the impact of a variety of health conditions. Diseases such as Ratoon Stunting Disease (RSD), Orange Rust, and Sugarcane Yellow Leaf Virus (SCYLV) can reduce yield by between 43\% and 50\% in severe circumstances \citep{RN8, RN9, RN10, RN11}. A study by \citet{RN12}, suggested that in 2019, there was an annual economic loss of \$25 million as a result of RSD infection with varying rates of incident across 87000 ha of sugarcane in Australia. Other factors that diminish crop health and contribute to a reduction in sugarcane yield include parasites, crop stress, and irrigation problems (for irrigated cane). Implementing methods to easily diagnose and early detection of any of these conditions on a large scale is vital for improving yield and the overall farming economics.

There are several methods the agricultural industry currently utilises to determine existing health conditions. The key motivation behind these health monitoring technologies is to increase the likelihood of early detection, preventing the condition from spreading or further deterioration in currently affected crops. A majority of current disease detection methods require visual inspection of symptoms or laboratory testing to conclusively determine the presence of a disease \citep{RN13, RN477, RN475, RN478, RN476}. Although accurate, collecting and testing enough samples to adequately evaluate the health of the crops is time-consuming and infeasible in large-scale agriculture. For example, the average sugarcane farm in Queensland is 110 Ha as of 2019 \citep{RN14} which makes visual inspection and sample collection challenging. Although many sugarcane diseases are identifiable through the morphological changes of a crop, for some diseases, the crops may be asymptomatic until the later stages of their infection. In such instances, visual inspection may no longer be reliable \citep{RN15}. Furthermore, sugarcane canopy can reach 2 to 7 meters, and very dense vegetation, prevents personnel from traversing through fields of mature sugarcane hence visual inspection can be difficult for matured crops \citep{RN16, RN17}. This makes monitoring the crop for health conditions such as parasites, stress, and irrigation issues, difficult, if not impossible. Thus, it would be beneficial to implement large-scale remote sensing for crop health monitoring that could identify crop health hazards before widespread impact and/or visible symptoms.

Several studies have shown the possibility of distinguishing healthy and unhealthy sugarcane without traditional laboratory testing \citep{RN18, RN92, RN15, RN20, RN224, RN225, RN95, RN213, RN93}. These studies focused on determining if there is a distinctive difference in the spectral reflectance of healthy and unhealthy crops with field spectroscopy for a variety of diseases. There was a 73\% and 96.2\% accuracy in identifying SCYLV and Orange Rust disease in sugarcane with hyperspectral imaging \citep{RN15, RN18, RN19}. In addition to sugarcane, multispectral imaging has been seen to be an effective disease detection method in other crops, with an 88\% accuracy in detecting several diseases in wheat, which like sugarcane is a perennial grass \citep{RN20}. Spectroscopy provides an opportunity to be able to diagnose health conditions at the plantations, which may not be visible to the human eye. 

This review intends to critically assess the current literature regarding health condition monitoring in sugarcane with the use of satellite spectroscopy. Additionally, we aim to collate, synthesise, and compare relevant literature on influential factors pertaining to the possible development of a large-scale health monitoring system for sugarcane with freely available satellite-based spectroscopy. A summary of the collated research is shown in Table ~\ref{tab3}, with the inclusion of examples of two other crops that are relevant to the research objectives of this review. 
Fig.~\ref{fig14} demonstrates the scope of our study and the qualitative relationship between the topics that we cover pertinent to the development of a spectroscopy-enabled sugarcane health monitoring system. 

Gaps in the literature regarding a system of this nature can then be identified and discussed with the overall objective of increasing the prevalence of satellite-sensing-based precision agriculture in the sugarcane industry to increase yield. Influential factors identified and discussed include:

\begin{enumerate}
    \itemsep0em
    \item Relevant information regarding sugarcane disease, growth limiting factors, optimal growth conditions and traditional disease detection methods.
    \item The most appropriate freely available satellites.
    \item Relevant vegetation indices.
    \item Factors that influence the observed reflectance of sugarcane.
    \item Utilized machine learning methodology and approach.
    
\end{enumerate}

\section{Literature Search Methodology}
Influential factors pertaining to the possible development of a sugarcane health monitoring system with satellite-based spectroscopy, including from free satellites, were identified through the review process of current literature and systems of this nature. Google Scholar only returned 15 relevant papers published before January 2024 on systems of this nature . The Keywords initially utilized to identify


\onecolumn
\begin{landscape}

\fontsize{8}{8}\selectfont
\begin{longtable}{p{3cm}p{2cm}p{1.5cm}p{2cm}p{3cm} p{1.5cm}p{1.5cm}p{1.5cm}p{2cm}}
\caption{Current multispectral and hyperspectral monitoring studies relevant to the development of a large scale sugarcane health monitoring system.\label{tab3}} \\
\hline\hline
		\textbf{Reference}	& \textbf{Health Condition}	& \textbf{Crop}	& \textbf{Spectroscopy} & \textbf{Category} &        \textbf{Satellite Name} & \textbf{Time Series} & \textbf{Machine Learning Algorithm} & \textbf{Best Classification Accuracy} \\
\hline\endhead  
\hline\endfoot  

    \citet{RN18, RN19} & Orange Rust & Sugarcane & Hyperspectral & Satellite & EO-1 Hyperion & False & LDA & 96.90\% \\
    \midrule
    \citet{RN92} & Mosaic & Sugarcane & Hyperspectral & Drone & N/A & False & SID & 92.50\% \\
    \midrule
    \citet{RN15} & SCYLV & Sugarcane & Hyperspectral & Handheld Spectrometer & N/A & True & DA & 73\% \\
    \midrule
    \citet{RN20} & White Leaf Disease & Sugarcane & Multispectral & Drone & N/A & False & RF, DT, KNN, XGB & 92\%, 91\%, 92\%, 92\% \\
    \midrule
    \citet{RN224} & Orange \& Brown Rust & Sugarcane & Multispectral & Drone & N/A & False & RF, KNN, SVM KNN & 90\%, 90\%, 90\%; 86\%, 83\%, 88\%\\
    \midrule
    \citet{RN225} & Brown Stripe \& Ring Spot & Sugarcane & Hyperspectral & Handheld Spectrometer & N/A & False &  RF, SVM, NB & 95\%, 85\%, 77\% \\
    \midrule
    \citet{RN94} & Powdery Mildew \& leaf rust & Wheat & Multispectral & Satellite & QuickBird & True & DT & 56.8\%, 65.9\%, 88.6\% \\
    \midrule
    \citet{RN95} & Cane Grub & Sugarcane & Multispectral & Satellite & Geo-Eye1 & False & GEOBIA & 79\% \\
    \midrule
    \citet{RN213} & Cane Grub & Sugarcane & Multispectral & Satellite & Geo-Eye1 & False & GEOBIA & 98.7\% \\
    \midrule
    \citet{RN72} & Non-specific & Mustard & Hyperspectral \& Multispectral & Satellite & EO-1 Hyperion \& LISS-IV & False & DT & N/A \\
    \midrule
    \citet{RN479} & Smut & Sugarcane & Hyperspectral & Handheld Spectrometer & N/A & False & CNN & \%90.9 \\
    \midrule
    \citet{RN93} & Diatraea saccharalis & Sugarcane & Hyperspectral \& Multispectral & Handheld Spectrometer \& Satellite & Landsat 8 & False & N/A & 79.8\% \& 85.5\% \\
    \midrule
    \citet{RN96} & Thrips & Sugarcane & Hyperspectral & Handheld Spectrometer & N/A & True & N/A & N/A \\
    \midrule
    \citet{RN216} & Orange \& Brown Rust & Sugarcane & Hyperspectral \& Multispectral & Handheld Spectrometer \& Drone & N/A & False & N/A & N/A \\

\end{longtable}
\end{landscape}

\begin{figure}
	\centering
        \includegraphics[width=5cm]{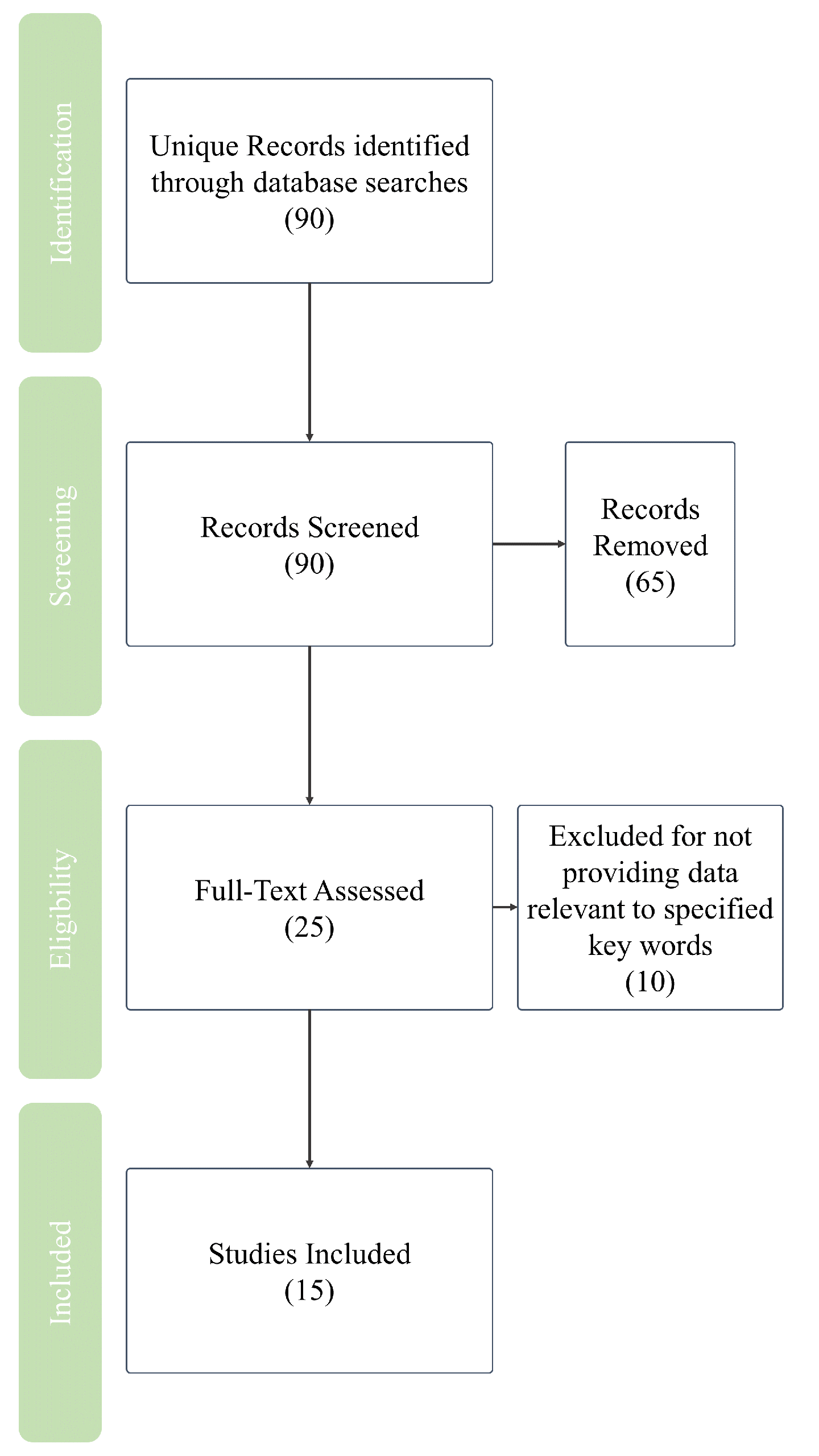}
        \caption{Literature Review flow diagram. \label{fig2}}
        
\end{figure}

\twocolumn

 \noindent literature were "sugarcane", "disease", "remote sensing", "pest", "health monitoring", "spectroscopy", "vegetation index", "reflectance" and "satellite". This search only evaluated literature to be relevant if it contained at least five of the nine specified keywords. Our literature search methodology is visualised in Fig.~\ref{fig2}.

\section{Sugarcane Health Conditions \& Growth Limiting Factors}
\subsection{Sugarcane diseases}
 Disease represents a significant challenge in sugarcane cultivation, prompting studies to explore their underlying mechanisms to aid in the development of more sophisticated detection methods. Orange rust is a fungal disease that produces orange patchy lesions from which moisture escapes, provoking a change in leaf structure \citep{RN36}.
 
 Infection typically occurs in wet humid conditions and is propagated by water, resulting in yield losses of up to 38\% \citep{RN9}. Brown rust bears a resemblance to orange rust with the variation between them suggested in their names. Yield losses of up to 22\% have been observed as a result of brown rust infection which typically occurs in dryer climates. Similar to the preceding health conditions, the Sugarcane Mosaic Virus has visible symptoms present on the foliage. This virus can reduce yield by 10\% to 50\% depending on the susceptibility of the variety and its region of origin \citep{ RN209, RN210, RN211, RN212}. Although these diseases can cause significant yield loss, a large number of current commercial sugarcane varieties in Australia are resistant to Orange rust, Brown rust and Mosaic virus \citep{RN34}. Significant losses can be incurred from infection of White Leaf disease with yield losses of at least 50\% previously observed by \citet{RN221} and can be visually identified from the striking white foliage. \citet{RN301} observed that Brown Stripe which is often a result of poor soil nutrition can result in yield losses of up to 40\% and \citep{RN36}. SCYLV has been observed to cause up to 37\% yield loss \citet{RN8}. Despite visual discoloration, Ring spot causes no reduction in yield and has no recommended treatment as a result \citep{RN36}. The resistance of Australian sugarcane varieties to Ring Spot, Brown Stripe, White Lead Disease and SCYLV are not known \citep{RN34}.

Unlike the preceding conditions, some sugarcane diseases such as RSD do not typically exhibit easily identifiable symptoms \citep{RN3, RN36}. RSD is caused by a bacteria, \emph{Leifsonia xyli}, which infects the xylem vessels of the sugarcane \citep{RN3, RN474}. It has been indicated that sugarcane in some regions is significantly more susceptible to RSD than other known diseases \citep{RN34}. In order to officially diagnose RSD, one of the following tests must be performed; evaporative-binding enzyme-linked immunoassay (EB-ELISA) or polymerase chain reaction (PCR). These tests are time- and cost-prohibitive, and are difficult to perform on large scales \citep{RN13, RN3}. The development of a large-scale continuous monitoring and detection method would, therefore, increase the detection of disease, helping to prevent further spread.

A practical solution to a large-scale sugarcane health monitoring system will need to be capable of identifying health conditions simultaneously. However, currently, only one study has attempted to classify multiple conditions  \citep{RN225}. Spectroscopy has been most commonly utilized in studies that consider Orange Rust and Brown Rust \citep{RN216, RN18, RN19, RN224}. This is done even though a majority of commercially used varieties are sufficiently resistant to the disease \citep{RN34}. It was observed that the greatest variation in the electromagnetic spectrum for both diseases occurred in the visible and NIR regions \citep{RN216, RN18, RN19}. These studies utilized either hyperspectral or multispectral imaging, and at different scales with a combination of Satellites, drones and handheld spectrometers employed. Where machine learning was implemented, orange rust was more accurately detected than any other disease with classification accuracies between 90\% and 96.9\%. 

While Orange Rust has been more accurately classified in the existing literature, there are certain factors inherent in the methodology of this classification that potentially contribute to this trend. Although the study by \citet{RN224} maintained a consistent methodology, it primarily focused on classifying varieties resistant to Orange and Brown rust, rather than identifying infected vegetation. Consequently, additional research is necessary to ascertain which regions of the electromagnetic spectrum exhibit the most significant variation due to specific symptoms and to validate that the specified diseases can be reliably distinguished from healthy sugarcane.

\subsection{Sugarcane pests}
Similar to diseases, pests contribute to a large reduction in the yield of sugarcane around the world. A prominent pest in Central and South America is a species of moth, Diatraea saccharalis, whose larvae bore into the internodes resulting in a yield reduction of between 13.5\% and 21.0\% \citep{RN203}. This is similar to the 13\% yield reduction observed across the industry between 1935 and 1957 by \citet{RN202} and in recent years a simple conservative estimation of yield loss was generalised to 0.77\% per 1\% of damaged internodes \citep{RN201}. Thrips pose a similar threat to sugarcane in Southeast Asia, South Africa and Australia by eating the sugarcane leaves, causing lacerations and discolouration resulting in yield reductions of between 18.0\% and 26.8\% \citep{RN204, RN205}. 
Contrary to the aforementioned pests, Canegrubs feed on the root system of sugarcane which results in reduced growth, lodging or plant death, significantly affecting the sugarcane industry. For instance, this pest could cost the Australian sugarcane industry an estimated \$40 million during severe outbreaks \citep{RN206, RN207}. 

The field of sugarcane pest detection has seen significant advancements, with various studies employing diverse methodologies and technologies. Notably, Canegrub infestations have been a focal point of research with recent studies \citep{RN95, RN213} investigating the infestations with multispectral imagery from the Geo-Eye1 satellite, employing Geographic Object-Based Image Analysis (GEOBIA) achieving classification accuracies of between 75\% and 98.7\%. Unlike diseases that may exhibit spectral variations related to moisture content, the symptoms of Canegrub infestations, often leading to lodging and significant soil exposure, influence the visible and near-infrared (NIR) spectra more significantly. Consequently, the detection of Canegrub infestations appears to be less dependent on water absorption bands and may contribute to the high classification accuracy of Canegrub symptoms whilst being constrained to wavelengths smaller than 1000nm. 

\citet{RN93} focused on Diatraea saccharalis, utilizing hyperspectral and multispectral data from Landsat 8 and a handheld spectrometer. Their approach, achieving classification accuracies of 79.8\% and 85.5\%, highlights the significance of incorporating various sensing technologies. However, the methodology and the explanation lack clarity. In contrast, \citet{RN96} explored Thrips detection using hyperspectral data and a handheld spectrometer. While specific accuracy metrics are not provided, the study emphasizes the potential of hyperspectral imaging for pest identification. 

Despite advancements, the field faces challenges similar to those in disease detection, including a lack of standardization and consistency among studies. Further research is warranted to identify optimal spectral regions for discerning specific pest-induced symptoms and to enhance the reliability of pest detection methodologies in sugarcane cultivation. Additionally, the development of a large-scale continuous monitoring and detection method would increase the detection of these pests, providing the opportunity to implement targeted control measures. Therefore, in this study, we focus on the conditions that have been detected using satellite-based spectroscopy where possible, in particular hyperspectral and multispectral images. 

\section{Spectral Imagery}
Hyperspectral and multispectral imaging are two forms of spectroscopy, which is the process of capturing images of reflected electromagnetic radiation from a number of bands across the electromagnetic spectrum, both the visible and non-visible light spectrum \citep{RN55, RN56}. A band is described as a small section of the electromagnetic spectrum. Each image captured is three-dimensional, consisting of an x and y spatial dimension, and a spectral dimension. The spatial resolution is defined by the physical dimensions of each pixel, while the spectral resolution is characterised by the number and size of the bands that are able to be captured. Hyperspectral imaging is a continuous set of narrow bands within the range of 400nm to 2500nm with intervals typically less than 10nm \citep{RN55, RN56, RN57}. This provides the opportunity to form a high-resolution continuous spectral signature. In contrast, multispectral imaging captures the spectral information for a discrete number of bands within the range of 400nm to 2500nm. Generally, multispectral imagery has fewer bands with varying bandwidths between 20 and 200nm \citep{RN58}. This results in multispectral imaging having a much lower spectral resolution than hyperspectral imaging. Although a higher spectral resolution is desirable, it comes at the cost of increased complexity and computational requirements. Additionally, there are significantly fewer satellites available with hyperspectral imaging capabilities than multispectral imaging. It is important to note that both hyperspectral and multispectral imaging can be captured at different spatial scales, with small-scale observations often acquired with drones or handheld spectrometers, while large-scale assessments are conducted through satellite platforms. This distinction in spatial scales introduces additional considerations when interpreting and comparing findings across studies in the field of sugarcane health condition detection.

\citet{RN18, RN19} found that Orange Rust was more accurately classified when considering the moisture content of the vegetation which can be indicated by wavelengths of 850, 1250, 1400 and 1650nm \citep{RN79, RN18, RN78, RN302, RN35}. However, typically in studies that demonstrated lower rates of classification, the spectral reflectance measurements were limited to 1000nm or less \citep{RN15, RN94, RN95, RN93}. In contrast, sensors with a wider spectral range typically exhibited higher classification rates \citep{RN224, RN225, RN19}. This is not a strict rule given that high classification rates were still seen with a limited spectral range \citep{RN92, RN20}. However, it is essential to note that the observed high rates of classification in these studies may be attributed to the specific characteristics of the diseases investigated. For instance, the diseases examined in these studies are known to cause white or significantly light discoloration in the leaves. The distinctive changes in the visible region of the electromagnetic spectrum could have been pronounced enough to achieve accurate classification without the need for additional information, thereby reducing the reliance on information from wavelengths beyond 1000nm. While studies focusing on diseases with different visual symptoms may necessitate a broader spectral range for effective classification, the particularities of the diseases considered in the cited research may explain success with limited spectral measurements. 

\subsection{Large Scale Spectroscopy}
\subsubsection{Hyperspectral}
The only study to date which classifies diseased sugarcane with satellite-based hyperspectral imaging was conducted by \citep{RN19}. The study effectively addresses noise and atmospheric effects through standard techniques of its time. This demonstrates potential for automation in developing an efficient health monitoring system after updating to more recent techniques and scaling the approach. The study found distinct spectral reflectance variations between healthy and unhealthy sugarcane across the electromagnetic spectrum. The high spectral resolution of hyperspectral imagery offers the advantage of capturing nuanced differences specific to particular health conditions, which may make it preferable when developing sophisticated models that differentiate multiple conditions. However, limitations due to data storage and processing power at the specified satellite sampling frequency may necessitate compromising accuracy. Multispectral imaging may be an effective compromise for these constraints, but more research is required to adequately conclude the required or ideal spectral resolution for large scale sugarcane health monitoring systems. Furthermore, additional investigations are essential to ascertain the reproducibility of these findings for various health conditions, sugarcane varieties, across diverse regions and under varying meteorological conditions preceding the acquisition of satellite imagery. The implications of these investigations significantly influence the feasibility of establishing a robust health monitoring system that is financially viable. 

\subsubsection{Multispectral}
Sugarcane health condition monitoring with satellite-based multispectral imaging has seen more attention than its hyperspectral counterpart \citep{RN93, RN95, RN213}. These studies all focus on pests and there are currently no studies that demonstrate the efficacy of satellite-based multispectral imaging for discerning various sugarcane diseases. Consequently, there is a lack of assessments against hyperspectral imagery counterparts with established disease detection methods. Additionally, the optimal number of bands and their respective bandwidths for multispectral satellites has yet to be determined in the context of differentiating sugarcane health conditions. Although \citet{RN64} compared the effect of bandwidth, they predominantly utilized satellites that are no longer in operation. Therefore, there is a prevailing need to revisit and refine this investigation using contemporary satellites.

The health condition monitoring study by \citet{RN95}, mapped canegrub damage using satellite-based multispectral imaging and achieved classification accuracies of between 53\% - 79\%. They further improved the accuracy of canegrub damage detection to between 75\% and 98.7\% through the implementation of a spectral difference segmentation algorithm into their established method \citet{RN213}. Both models implemented a Geographic Object-Based Image Analysis (GEOBIA) approach to consider contextual and shape information in the analysis. Both studies derived the NDVI to indicate vegetation health and assessed the standard deviation of the red spectral band to indicate the texture. Empirically derived thresholds were established for the 30-quantile of NDVI and the 70-quantile of the red spectral band standard deviation. Regions falling below the 30-quantile and surpassing the 70-quantile were identified as potential areas affected by canegrub damage. The severity of the damage was classified using a set of undisclosed thresholds based on the relative differences in NDVI and texture values compared to the rest of the block, however, the rationale behind determining these thresholds was not included in the paper. Additionally, there was little information on the effectiveness of the models' ability to distinguish between canegrub damage and analogous symptoms such as lodging.  

\citet{RN66} performed sugarcane variety classification in Brazil using Sentinel-2 image and found that multispectral satellites can be utilized to perform classification of sugarcane based on variations in spectral reflectance. Although there are only a few studies on using satellite-based multispectral imaging in sugarcane health monitoring, it has been used in several studies monitoring the health of other crops. For example, the study by \citet{RN72} incorporated multispectral imaging from LISS-IV in image fusion where images of high spectral resolution are fused with images of high spatial resolution to estimate an image of both high spatial and spectral resolution \citep{RN97, RN98}. However, the complexity of this approach hinders a direct assessment of the sustainability of satellite-based multispectral imagery for a large-scale health monitoring system for sugarcane. Consequently, the lack of clarity on the minimum required spatial resolution for effective health monitoring and the potential impact of atmospheric conditions on the viability of satellite-based imagery for this purpose underscores the need for further studies in this domain. Once a comprehensive understanding of the impacts associated with the choice of sensor, in the context of sugarcane health monitoring, is established, the implementation of image fusion should be explored further. The ability to combine the strengths of high-spectral-resolution and high-spatial-resolution images, positions image fusion as a powerful technique with the potential to enhance the overall effectiveness of monitoring systems. This approach, when employed judiciously, can contribute to more accurate and detailed insights into sugarcane health, paving the way for advanced disease detection methodologies, once the fundamentals have been sufficiently understood.

In another example of using multispectral satellite for health monitoring,\citet{RN94} found multispectral imaging could be utilized to detect powdery mildew and leaf rust in wheat. This indicates it may be plausible to adopt a similar approach in attempting to identify sugarcane-based health conditions with satellite-based multispectral imaging. However, the viability of a multispectral satellite for disease detection will vary for each satellite as each sensor will capture different bands with varying bandwidths and resolutions. \citet{RN94} utilized the commercial high-resolution multispectral satellite QuickBird with a resolution of 2.4m \citep{RN99} and therefore a similar approach with a lower resolution freely available satellite for sugarcane may not yield as effective classification. Therefore, before attempting to determine the optimal characteristics of multispectral satellites for disease detection, the viability of conducting disease detection for sugarcane at all with the currently available free multispectral satellites should be investigated. The methodology required to perform disease detection with satellite-based multispectral imagery will remain consistent with the methodology to perform satellite-based hyperspectral disease detection, with variations in the vegetation indices and pre-processing methods utilized. 

\subsection{Small Scale Spectroscopy}
The majority of previous studies for sugarcane were performed with field spectroscopy or drone rather than satellite \citep{RN92, RN15, RN20, RN224, RN225, RN96, RN216}. The impracticality of field spectroscopy in sugarcane plantations stems from the crop's extensive scale and dense vegetation, rendering the approach comparable to traditional sampling methods without significant gains in detection capabilities. This comparison is valid, as a considerable number of diseases affecting sugarcane can be reliably diagnosed through visible inspections conducted by trained personnel \citep{RN23, RN36}. While employing field spectroscopy may enhance confidence in diagnosis, it does not enhance the efficacy of performing widespread disease detection any more than employing trained field agronomists. The heart of the issue is that only sugarcane on the periphery of sugarcane plantations can be visually inspected. By the time the disease had spread to the visual periphery of the plantations, it would have likely become a widespread issue. 

The majority of previous studies on sugarcane disease detection conducted in the last 5 years have favored the use of drones \citep{RN92, RN20, RN224, RN216}. This trend suggests a shift towards approaches that offer more flexibility in terms of accessibility and manoeuvrability. The agility of disease detection tools is paramount for effective monitoring in the dynamic and expansive context of sugarcane plantations. Although drones can perform these tasks on a wider scale when compared with hand-held devices, typically drones are not capable of operating over distances as large as the standard sugarcane plantation and may require several flights. Furthermore, there are often stringent legal requirements for operating drones and costs associated with the acquisition of the drone.  With that being said, it is essential to note that recent literature employing drones reported high classification results or statistically significant differences between healthy and diseased crops \citep{RN92, RN224, RN20, RN216}. This is not unexpected considering the potential access to high spatial and spectral resolution spectroscopy when utilizing drones. Satellite-based remote sensing may provide an effective alternative or complementary solution for crop health monitoring at a sufficiently large scale, provided it is able to produce consistent comparable classification results as drone-based monitoring. A future study should investigate trade-offs in classification accuracy and the logistics of each method.


\section{Vegetation Indices} \label{sec:vegind}

\begin{figure*}[H]
\includegraphics[width=\textwidth]{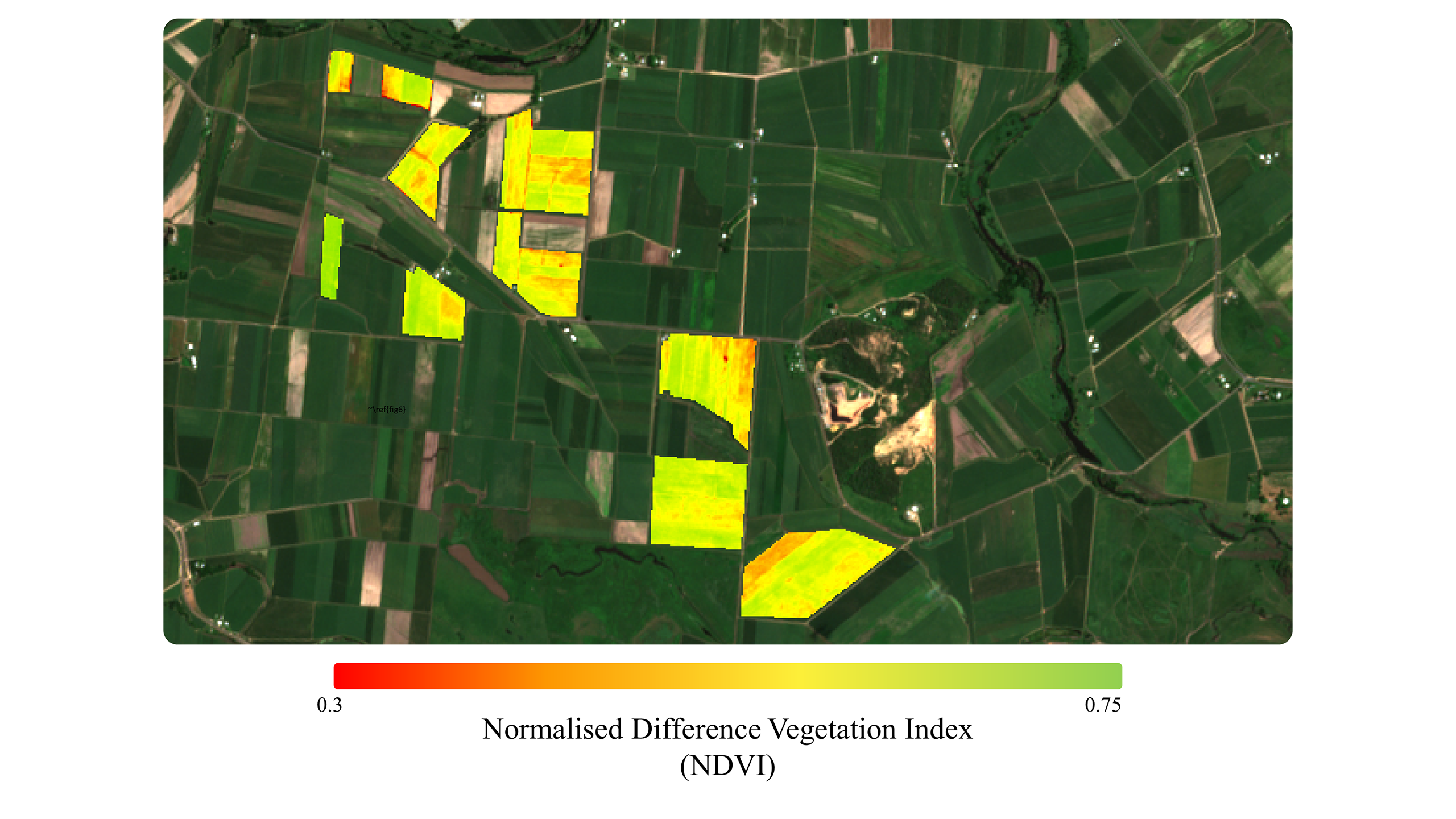}
\caption{Sentinel-2 10m spatial resolution true colour image of sugarcane farms in the Herbert region of Queensland, Australia, overlaid with an NDVI raster.  \label{fig10}}
\end{figure*}   

Hyperspectral and multispectral imaging offer an alternative to traditional health monitoring by detecting changes in sugarcane spectral reflectance as a result of the health condition. These changes in spectral reflectance occur as a consequence of phenotypic and morphological changes incurred from the presence of disease or other conditions. A transformation to spectral imagery known as a vegetation index can be performed to evaluate the state of vegetation for a variety of applications \citep{RN45}. Vegetation indices aim to accentuate important characteristics of the vegetation and reduce the impact of redundant factors \citep{RN45, RN71}. The result of the vegetation indices can then be utilized in algorithms to solve agriculture-related problems \citep{RN227, RN228}. The applicability of a given vegetation index to an application is dependent on the sensor utilized to capture the spectral reflectance of the vegetation and the specific objective of the project \citep{RN45}. Vegetation indices differ for hyperspectral and multispectral imaging as more specific bands with narrower bandwidths can be employed with hyperspectral imaging \citep{RN45}. One of the most commonly employed vegetation indices in precision agriculture is the Normalized Difference Vegetation Index (NDVI) \citep{RN45} which is utilized to indicate vegetation density and health. See for an example, Figure~\ref{fig10} which overlays an NDVI raster on a true colour image of sugarcane farms in the Herbert region of Queensland, Australia. This visually indicates variation between and within paddocks of sugarcane in the region, facilitating further investigation.

\citet{RN18, RN19} found success in developing five new vegetation indices, that consider the moisture content of vegetation and were dubbed the Disease Stress Water Indices (DSWI). One of these indices was in the linear combination that achieved the highest overall accuracy of 96.9\% for the classification of orange rust with LDA. Three out of the five DSWI had the highest canonical correlation with sugarcane infected with Orange Rust. These DSWI were formulated through the combination of Visible and Near-infrared (VNIR) and Short-Wave infrared (SWIR) reflectance bands. This was likely effective for orange rust given that it is a fungal disease that produces orange patchy lesions from which moisture escapes, provoking a change in leaf structure, and causing plant stress and moisture content, which corresponds with a change in the VNIR and SWIR reflectance, respectively \citep{RN18, RN19, RN78, RN302, RN35}. Therefore, it is reasonable to expect these indices to be effective for the detection of other diseases that alter the leaf structure or moisture content of vegetation \citep{RN72}. The DSWI may not be effective for diseases that do not alter leaf structure, plant stress, and moisture content. For example, in the case of RSD, where the leaf structure of sugarcane remains the same and just the absorbance of water is affected \citep{RN36}, the DSWI may not respond as effectively as a typical disease. This presents an opportunity to potentially utilise additional vegetation indices which focus on the moisture content of the vegetation \citep{RN78, RN302} or are combined with the DSWI \citep{RN45}.  

The prevailing trend in current literature predominantly focuses on the assessment of raw spectral measurements and at most a single individual vegetation index, with many studies neglecting to consider more than one \citep{RN15, RN94, RN96, RN216, RN72, RN93, RN95, RN213, RN92, RN225}. This is despite the existence of several studies that assessed and showed that multiple vegetation indices consistently demonstrated high classification accuracies \citep{RN19, RN20, RN224}. Previous success in classifying Orange Rust in sugarcane was found based on vegetation indices that accentuate variation in the moisture content of vegetation, thereby indicating that other moisture-based vegetation indices should be heavily considered in future studies. Table~\ref{tab1} contains vegetation indices from the collated literature which focus on changes in moisture content, leaf pigments, and leaf structure for plausible use in a sugarcane health monitoring system. 

\onecolumn
\begin{longtable}{>{\centering\arraybackslash}p{0.25\textwidth}>{\centering\arraybackslash}p{0.35\textwidth}>{\centering\arraybackslash}p{0.35\textwidth}}

\caption{Vegetation indices with a plausible application for sugarcane health monitoring.\label{tab1}} \\
\toprule
\textbf{Reference}	& \textbf{Vegetation Index}	& \textbf{Formula}\\

\midrule
\citet{RN73} & Normalized Difference Vegetation Index (NDVI) & $\frac{NIR - RED}{NIR + RED}$\\
\midrule
\citet{RN74} & Atmospherically Resistant Vegetation Index (ARVI) & $\frac{NIR - (RED-BLUE)}{NIR + (RED-BLUE)}$\\
\midrule
\citet{RN75} & Simple ratio index (SRI) & $\frac{NIR}{RED}$\\
\midrule
\citet{RN76} & Plant Senescence Reflectance index (PSRI) & $\frac{R_{680} - R_{500}}{R_{750}}$\\
\midrule
\cite{RN77} & Ratio Vegetation Index (RVI) & $\frac{RED}{NIR}$\\
\midrule
\citet{RN78} & Normalized difference Water index (NDWI)	& $\frac{GREEN - NIR}{GREEN + NIR}$\\
\midrule
\citet{RN79} & Normalized Difference Moiseture Index(NDMI)	& $\frac{NIR - SWIR}{NIR + SWIR}$\\
\midrule
\citet{RN80} & Normalized Green Red Difference Index NGRDI & $\frac{GREEN - RED}{GREEN + RED}$\\
\midrule
\citet{RN81} & Non-Linear Index (NLI) & $\frac{NIR^2 - RED}{NIR^2 + RED}$\\
\midrule
\citet{RN82} & Visible Atmospherically Resistant Index (VARI)	& $\frac{GREEN - RED}{GREEN+RED - BLUE}$\\
\midrule
\citet{RN217} &  Modified Chlorophyll Absorption in Reflectance Index (MCARI)	& $[(R_{700} - R_{670}) - (0.2(R_{700} - R_{550}))\frac{R_{700}}{R_{670}}]$\\
\midrule
\citet{RN218} &  Transformed Chlorophyll Absorption in Reflectance Index (TCARI) & $3[(R_{700} - R_{670}) - (0.2(R_{700} - R_{550}))\frac{R_{700}}{R_{670}}]$\\
\midrule
\citet{RN219} & Photochemical Reflectance Index (PRI) & $\frac{R_{531} - R_{570}}{R_{531} + R_{570}}$\\
\midrule
\citet{RN220} & Pigment Specific Simple Ratio (chlorophyll a) (PSSRa) & $\frac{R_{800}}{R_{680}}$\\
\midrule
\citet{RN220} &  Pigment Specific Simple Ratio (chlorophyll b) (PSSRb) & $\frac{R_{800}}{R_{635}}$\\
\midrule
\citet{RN18, RN19} & DSWI-1 & $\frac{R800}{R1660}$\\
\midrule
\citet{RN18, RN19} & DSWI-2 & $\frac{R1660}{R550}$\\
\midrule
\citet{RN18, RN19} & DSWI-3 & $\frac{R1660}{R680}$\\
\midrule
\citet{RN18, RN19} & DSWI-4 & $\frac{R550}{R680}$\\
\midrule
\citet{RN18, RN19} & DSWI-5 & $\frac{R800 + R550}{R1660 + R680}$\\
\midrule
\citet{RN83} & Structure\-Insensitive Pigment Index (SIPI) & $\frac{R_{800} - R_{445}}{R_{800} - R_{680}}$\\
\midrule
\citet{RN450} & Simple Ratio (SR) & $\frac{R_{750}}{R_{705}}$ \\
\midrule
 & SR 800/550 & $\frac{R_{800}}{R_{550}}$ \\
\midrule
  & Normalized Difference (ND) 750/660 & $\frac{R_{750}-R_{660}}{R_{750}+R_{660}}$ \\
\midrule
\citet{RN451} & ND 800/680 & $\frac{R_{800}-R_{680}}{R_{800}+R_{680}}$ \\
\midrule
\citet{RN450} & ND 750/705 & $\frac{R_{750}-R_{705}}{R_{750}+R_{705}}$ \\
\midrule
\citet{RN451} & Modified SR (MSR) & $\frac{R_{750}-R_{445}}{R_{705}-R_{445}}$ \\ 
\midrule
\citet{RN451} & Modified ND (MND) & $\frac{R_{750}-R_{445}}{R_{750}+R_{705}-2R_{445}}$ \\
\midrule
\citet{RN450} & SR 750/550 & $\frac{R_{750}}{R_{555}}$ \\
\midrule
  & Ave(750–850) & Average between $R_{750}$ and $R_{850}$ \\
\midrule
\citet{RN452} & Modified Chlorophyll Absorption in Reflectance Index (MCARI) & $[(R_{700}-R_{670})-0.2(R_{700}-R_{550})]\frac{R_{700}}{R_{670}}$\\
\midrule
\citet{RN454} & Transformed Chlorophyll Absorption in Reflectance Index (TCARI) & $3[(R_{700}-R_{670})-0.2(R_{700}-R_{550})]\frac{R_{700}}{R_{670}}$\\
\midrule
\citet{RN453} &  Optimized Soil-Adjusted Vegetation Index OSAVI & $(1+0.16)\frac{R_{800}-R_{670}}{R_{800}+R_{670}+0.16}$ \\
\midrule
\citet{RN454} & Ratio TCARI/OSAVI & $\frac{\text{TCARI}}{\text{OSAVI}}$ \\
\midrule
\citet{RN455} & SR 695/420 & $\frac{R_{695}}{R_{420}}$ \\
\midrule
\citet{RN455} & SR 695/760 & $\frac{R_{695}}{R_{760}}$ \\
\midrule
\citet{RN456} & Green Normalized Difference Vegetation Index (GNDVI) & $\frac{(NIR - \text{Green})}{(NIR + \text{Green})}$ \\
\midrule
\citet{RN450} & Normalized Difference Red-edge Index (NDRE) & $\frac{(NIR - \text{Red-edge})}{(NIR + \text{Red-edge})}$ \\
\midrule
\citet{RN457} & Chlorophyll index – Green (CiGreen) & $\frac{NIR}{\text{Green}} - 1$ \\
\midrule
\citet{RN457} & Chlorophyll index – Red-edge (CiRE) & $\frac{NIR}{\text{Red-edge}} - 1$ \\
\midrule
\citet{RN458} & Difference Vegetation Index (DVI) & $NIR - \text{Red-edge}$ \\
\midrule
\citet{RN459} & Enhanced Vegetation Index (EVI) & $2.5 \times \frac{(NIR - \text{Red})}{(NIR + 6 \times \text{Red} - 7.5 \times \text{Blue} + 1)}$ \\
\midrule
\citet{RN460} & Chlorophyll Vegetation Index (CVI) & $\frac{(NIR \times \text{Red})}{\text{Green}^2}$ \\
\midrule
\citet{RN461} & Triangular Vegetation Index (TVI) & $0.5 \times (120 \times (NIR - \text{Green}) - 200 \times (\text{Red} - \text{Green}))$ \\
\midrule
\citet{RN462} & Plant Senescence Reflectance Index (PSRI) & $\frac{(\text{Red} - \text{Green})}{\text{Red-edge}}$ \\
\midrule
\citet{RN463} & Blue Green Pigment Index (BGI) & $\frac{\text{Blue}}{\text{Green}}$ \\
\midrule
\citet{RN464} & Green Leaf Index (GLI) & $\frac{(\text{Green} - \text{Red}) + (\text{Green} - \text{Blue})}{(2 \times \text{Green} + \text{Red} + \text{Blue})}$ \\
\midrule
\citet{RN465} & Modified Soil Adjusted Vegetation Index (MSAVI) & $\frac{2 \text{NIR} + 1 - \sqrt{(2 \text{NIR} + 1)^2 - 8( \text{NIR} -  \text{RED})}}{2}$ \\
\midrule 
\bottomrule
\end{longtable}
\begin{multicols}{2}


\section{Influential Factors on Sugarcane Reflectance}
\subsection{Sugarcane Variety}
The spectral reflectance of vegetation depends on the plant's physical and biochemical characteristics, including Leaf Area Index (LAI), leaf orientation, biomass, soil background, cell structure, moisture content and chlorophyll content \citep{RN35, RN56, RN57, RN59, RN60, RN61, RN62}. Given that the varieties of sugarcane differ from one another, it is reasonable to expect their spectral reflectance would demonstrate a corresponding change. \citet{RN64} found a significant difference in spectral reflectance of several Brazilian sugarcane varieties. Using multiple discriminant analysis (MDA), \citet{RN64} were able to achieve 87.5\%  correct classification in identifing the variety \citep{RN63, RN64}. \citet{RN33} performed a similar study with varieties prominent in Australia during that time period, successfully discriminating sugarcane varieties with a number of ML techniques, among which random forest (RF) and support vector machine (SVM) were found to be the most effective with 87.5\% and 90\% per pixel classification accuracy. Both studies utilized the EO-1 Hyperion satellite and indicated that the spectral reflectance of different sugarcane varieties can vary substantially, with the regions of greatest separability seen in the Near Infrared (NIR) and Short-Wave Infrared (SWIR) regions (See Figure 2 in \citet{RN33}). \citet{RN66} corroborated these results by classifying varieties with a classification accuracy of 86\% and 90\% with Sentinel-2. 

To establish an effective real-time disease and health monitoring system using satellite-based spectral imaging, it is essential to consider variety as a crucial factor. Without accounting for variety-specific characteristics, the detection of variations in spectral reflectance due to health conditions across multiple sugarcane varieties becomes impractical. Despite these findings, only three health monitoring studies considered variety \citep{RN15, RN92, RN224}. Unsurprisingly, \citet{RN15} found that classification was more accurate when performed within a variety rather than across multiple varieties, as a consequence of spectral variation between varieties. It was demonstrated in \citep{RN15} that SCYLV affects varieties differently with chlorophyll b levels of HO 95-988 remaining unaffected, but chlorophyll b levels of LCP 85-384 were significant when infected with SCYLV. This makes it reasonable to expect that sugarcane diseases may affect varieties in different manners and to different extents, incurring different alterations in the spectral reflectance. Similarly, it was demonstrated in a study conducted by \citet{RN96} that spectral reflectance of sugarcane varieties may react differently to Thrips damage. However, both studies conducted only featured two varieties of sugarcane and therefore further research should be conducted that considers variety as a factor.

\subsection{Meteorological Effects}
The majority of research on large-scale sugarcane health monitoring was based on images at a single location at one point in time \citep{RN18, RN19, RN92, RN20, RN95, RN224, RN96}. However, this prevailing pattern implies that annual variations in weather, such as temperature, humidity, sunlight duration, wind patterns and quantity of preceding rain, are not considered. By extension, it is currently unknown how effective a given method of health monitoring is when applied to different regions. Consequently, it is currently unknown whether the corresponding variation in these factors is considerable enough to impact the ability to discern health conditions with ML and therefore requires further research. Recognizing this limitation is imperative in the context of developing long-term health monitoring systems, where a comprehensive understanding of environmental dynamics is crucial for accurate and reliable assessment of diseases.

\subsection{Temporal Morphological Changes \& Multi-temporal Analysis}
\begin{figure*}[H]
\includegraphics[width=\textwidth]{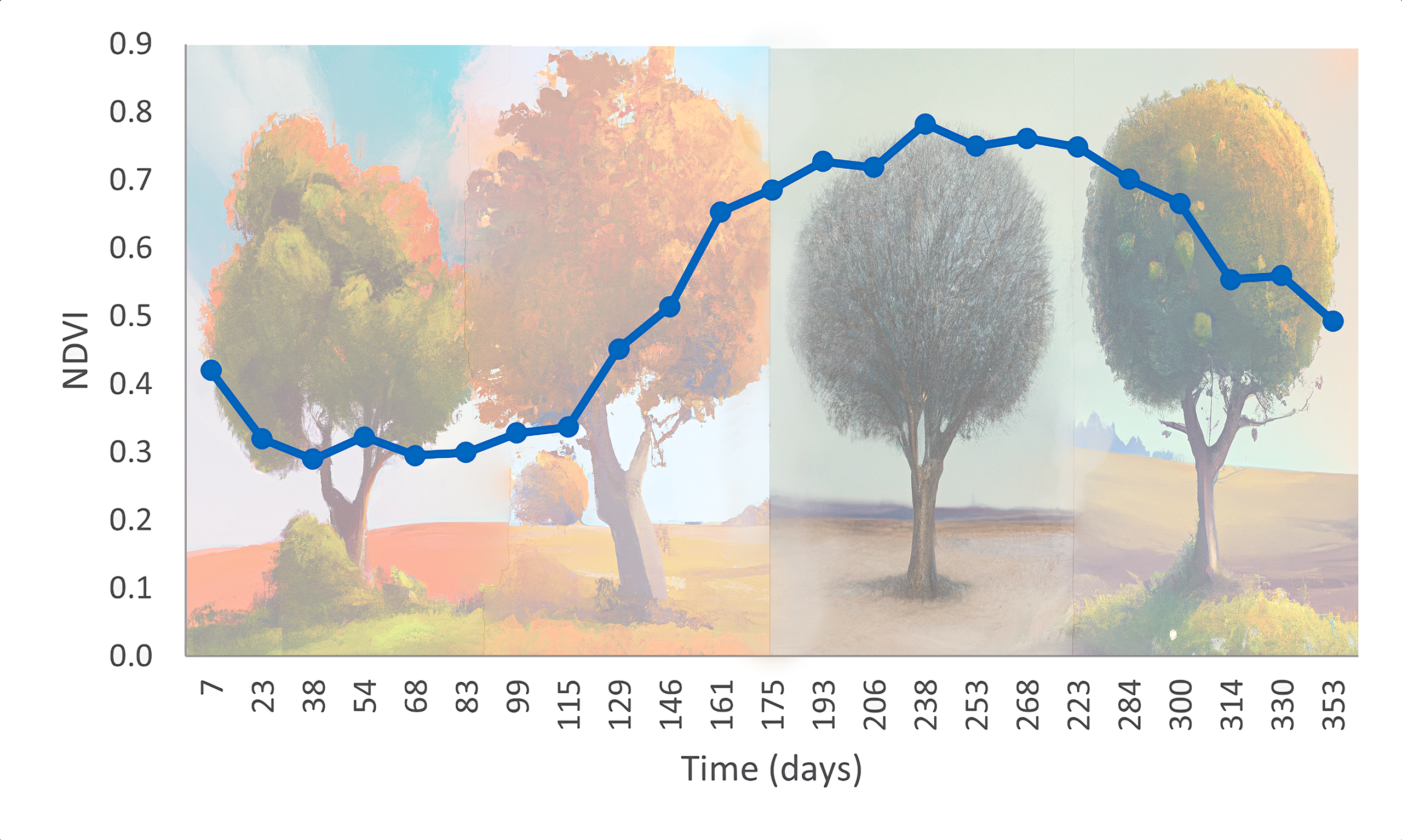}
\caption{NDVI measurements of sugarcane across entire growth cycle with MODIS and HJ-1 CCD remote sensing. Background images indicate approximate season. (Adapted from \citet{RN200} with Adobe Photoshop and AI generated artwork) \label{fig4}}
\end{figure*}

Growth and maturation cause phenotypic and morphological changes in characteristics that dictate the spectral reflectance in sugarcane. This was investigated and observed in 25 sugarcane fields in Thailand with imagery from Landsat-8 OLI \citep{RN16}. Additionally, the average NDVI of the fields was observed to change over time, where phases associated with the largest amount of growth exhibited a higher NDVI \citep{RN16}. A similar result was obtained and observed to vary across sugarcanes lifecycle by \citet{RN200} with MODIS and HJ-1 CCD remote sensing, as can be seen in Figur~\ref{fig4}. Comparable results were then achieved in Indonesia by \citet{RN215}. 

\citet{RN67, RN68} was able to distinguish phenological stages with multispectral imaging from Landsat 7, demonstrating that there is a significant difference in the spectral reflectance of sugarcane at different phenological stages. Additionally, temporal morphological changes indirectly contribute to the change in observed reflectance of sugarcane through flowering or lodging. This results from the reduction in the portion of canopy that is observable compared to soil, stalks and flowers, which exhibit different leaf orientation and cell structure \citep{RN30}. \citet{RN214} observed that the spectral reflectance of sugarcane varied at an annual scale and concluded it is likely attributed to a combination of the topography and rainfall. These factors may need to be considered to accurately classify variation in spectral refelctance as a consequence of health conditions rather than seasonal changes. 

The temporal aspect of sugarcane health monitoring has received limited research attention, with the majority of existing literature adopting a cross-sectional study approach \citep{RN18, RN19, RN92, RN20, RN95, RN224, RN96}. The viability of producing a real-time sugarcane health monitoring system hinges on being able to detect health conditions for sugarcane across its life cycle with multi-temporal data. An investigation was conducted into this for disease detection in wheat by \citet{RN94}. Three separate multispectral images were captured at varying points in a wheat paddock across its life cycle. Spectral mixture analysis was performed and a decision tree was utilized to classify disease at various points in the wheat’s life cycle with a classification accuracy of 56.8\%, 65.9\% and 88.6\%. Younger crops have significantly smaller biomass, and therefore the reflectance will be more heavily influenced by the background soil as the composition of any given pixel will incorporate a larger percentage of soil \citep{RN35, RN71, RN45}. Additionally, the machine learning algorithm employed in this study has been superseded by newer tree-based models; implementing the more recently developed models could yield improved outcomes for health monitoring. Further studies should be conducted incorporating in situ soil reflectance measurements and vegetation indices into the models to observe the impact on results for large-scale disease detection in crops at early life cycle stages. 

The study by \citet{RN15} on SCYLV was the only known study to conduct multi-temporal disease detection in sugarcane using field spectroscopy. The results of this study demonstrate variation in spectral reflectance as a plant matures \citep{RN16, RN35}. This difference in spectral reflectance was observed to decrease between the collection of the first samples on the 13th of July and the second group of samples collected on the 12th of October. It was then observed in the samples collected on November 4th that the difference in spectral reflectance increased to approximately the same as seen in the July samples. The observed fluctuations in spectral reflectance differences over time suggest that distinct stages of the sugarcane life cycle may exhibit varying reactions to a disease. This temporal variation in spectral characteristics could potentially impact the accuracy of disease classification, as noted by \citet{RN15}, emphasizing the importance of understanding and accounting for temporal changes in sugarcane health monitoring. It would be beneficial to conduct further research into multi-temporal disease detection with spectral data to consolidate these findings.

\subsection{Viewing Angle of Vegetation}
An additional factor to consider in the development of a satellite-based real-time health monitoring system is the effects of viewing angle on spectral reflectance. Sugarcane like many other crops is non-Lambertian, and therefore a change in the viewing angles will influence the spectral reflectance. \citet{RN69} investigated the effects of viewing angle variation in sugarcane radiometric measurements. They found a noticeable difference in the spectral reflectance profiles of sugarcane at different viewing angles (See Figure 3 in \citep{RN69}). The Bidirectional Reflectance Distribution Function (BRDF) correction model developed by \citet{RN70} can be applied to compensate for the anisotropy factor. Specifically, \citet{RN69} suggested that it could be utilized to correct hyperspectral images captured with an Unoccupied Aerial Vehicle (UAV) \citep{RN69}. The BRDF correction model was utilized to calculate the spectral reflectance at different viewing angles (See Figure 6 in \citep{RN69}). \citet{RN92} used the BRDF correction for sugarcane health monitoring and was the only health monitoring study which considered the effect of viewing angle. Future research should ensure spectral images are centred at a consistent location and viewing angle. Additionally, the BRDF correction model should be considered to handle the anisotropy factor. 

\section{Satellites for Health Monitoring}
Satellites have many applications in remote sensing, including but not limited to land surveying, earth science, and agriculture \citep{RN45, RN227, RN229, RN230}. It is highly advantageous to use satellite-based remote sensing for large or inaccessible areas, where data collection would be infeasible using conventional methods. Satellite-based remote sensing varies depending on the sensor. Typically in agricultural applications sensors include radiometers, spectrometers, Red Green Blue (RGB) cameras, panchromatic cameras, and thermal imaging cameras. There are many commercial and freely available satellites with a combination of the aforementioned sensors. Commercial satellites are often costly, which makes them unsuitable for small farms or wide use. To facilitate satellite use for affordable monitoring applications, in this study, we only focus on freely available satellites. A list of these satellites appropriate for precision agriculture can be seen in Table~\ref{tab2}. 

\subsection{Satellite selection considerations}
An important factor to consider is the temporal resolution of a satellite, specifically the period of time taken to return to the same position at nadir, repeating cycles and revisiting times. A repeat cycle is the time it takes for the satellite to be centred at the same previous latitude and longitude coordinates, at the same angle. Whereas the revisit time is the period of time before the same location is surveyed again at all. This resolution has been increased by some satellites through the deployment of two identical satellites 180 degrees out of phase. This is the case for the series of satellites Sentinel-2 and Sentinel-3. 

\begin{figure}
\includegraphics[width=\textwidth]{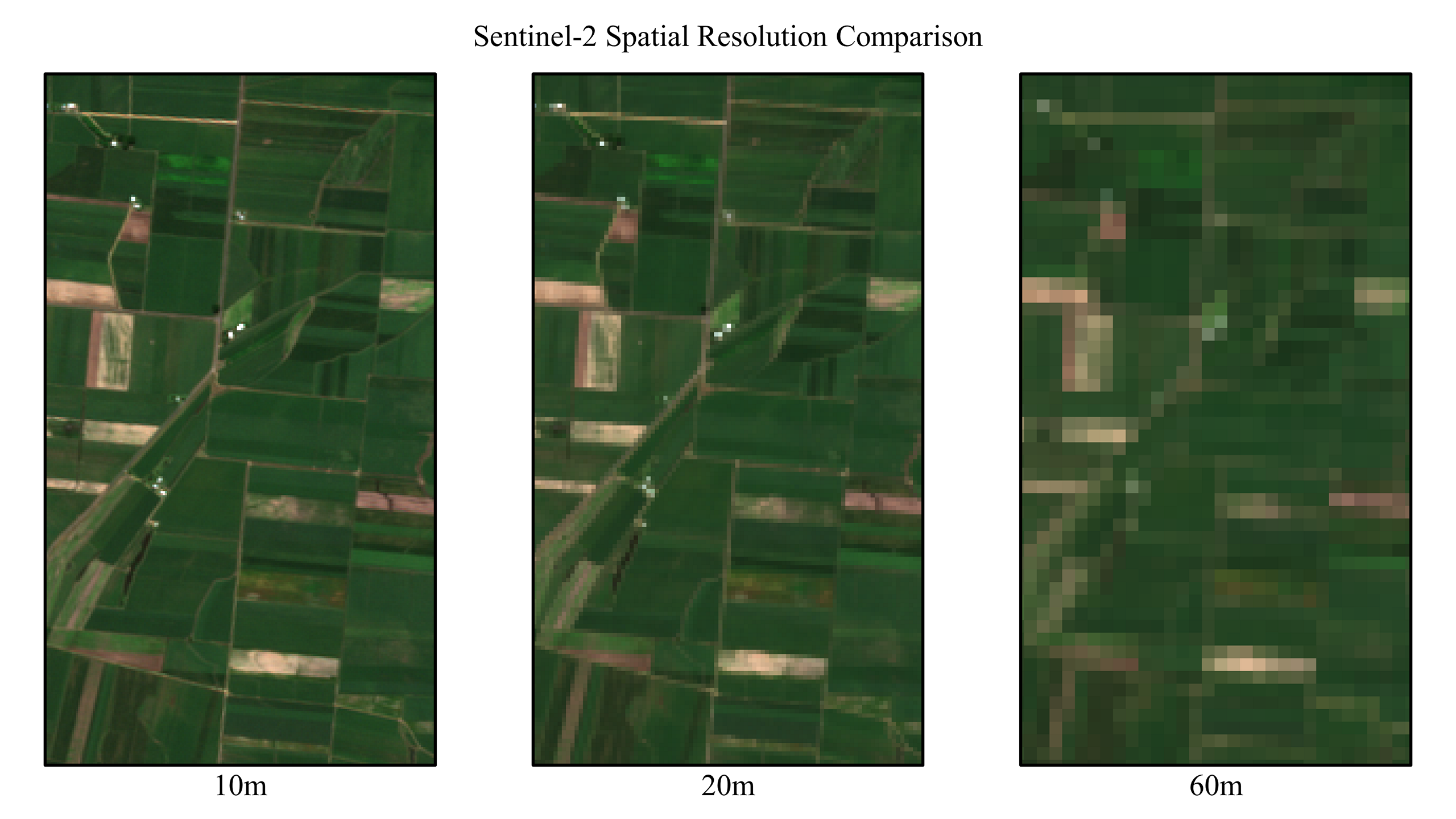}
\caption{Spatial resolution comparison of available Sentinel-2 resolutions. Sentinel-2 offers four bands at 10m resolution, ten bands at 20m resolution and twelve bands at 60m resolution\label{fig11}}
\end{figure}

Another major factor to consider when selecting an appropriate satellite for precision agriculture is the spatial resolution. This varies significantly between satellites and is often significantly better for commercial satellites. There is often a trade off between high spatial and spectral resolution, consequently, several freely available multispectral satellites offer varying spatial resolutions depending on the spectral bands available. Therefore, it is important to ensure satellites with the desired spectral bands are available in the required spatial resolution as the variation can be significant. For example, Figure~\ref{fig11} shows a visual comparison of a true colour image in the available Sentinel-2 spatial resolutions of 10m, 20m, and 60m. Sentinel-2 captures data across 13 different spectral bands between approximately 400nm and 2300nm, however only the red, green, and blue spectral bands are available at every resolution \citep{RN46, RN231}. There is only a single NIR band (Band 8) available with a 10m spatial resolution, centred at 842nm, in contrast to the four available in NIR region and two in the SWIR region at a spatial resolution of 20m \citep{RN46}. Therefore, monitoring specific wavelengths may limit the spatial resolution when considering financial constraints.

\subsection{Satellite Limitations}
Although satellites provide the ability for large-scale sensing, there are several limitations. Satellite imagery is susceptible to atmospheric effects and requires atmospheric correction to be applied to convert atmospheric radiance received by the satellite to an accurate surface reflection \citep{RN37, RN38}. These atmospheric effects vary depending on the spectral band and were found to produce a mean difference of 18\% between the NDVI of corrected and non-corrected values \citep{RN38, RN39}. Several techniques and programs can be utilized to perform atmospheric correction.

A review of approaches to atmospheric correction was undertaken by \citet{RN40} which included several commercially available software packages. Quick Atmospheric Correction (QUAC) and Fast Line-of-Sight Atmospheric Analysis of Spectral Hypercubes (FLAASH) are two modules available for the software, Environment for Visualising Images (ENVI), to perform atmospheric correction \citep{RN40, RN41, RN42}. Other software such as Atmospheric and Topographic Correction (ATCOR) is available to perform atmospheric correction for a variety of remote sensing applications \citep{RN40, RN41, RN43}. The Earth Resource Data Analysis System (ERDAS) and ENVI are software packages that can also perform a number of other remote sensing pre-processing tasks including destripping and image registration. The review conducted by \citet{RN40} highlighted that the performance of atmospheric correction techniques varied depending on the specific type of landscape in consideration. The Empirical Line Method (ELM) performed well \citep{RN40}, however it requires reflectance field measurements. Therefore, when evaluating atmospheric correction techniques for large-scale health monitoring, it is imperative to consider the landscape and logistical availability of spectrometers for quantitative calibrations. 

In addition to the above-mentioned commercial software, there are freely available packages capable of performing atmospheric correction. Python-Based Atmospheric Correction (PACO), is a Python library that is based on the ATCOR IDL code \citep{RN41}. The initial release of the software is currently operational for Sentinel-2 series, Landsat-8, DESIS, and EnMAP satellites and currently has uncertainty values of approximately 30\% and 10\% for retrieval aerosol optical thickness and water vapour, respectively. Alternatively, the darkest pixel is a common and simple image-based technique that could be implemented for other satellites not supported by the PACO library \citep{RN41}. A large body of water often corresponds to the darkest pixel within the scene. This technique assumes that the darkest pixel has a surface reflectance of approximately zero and that the majority of the reflectance is a result of scattering in the visible light spectrum \citep{RN38, RN44}.

\end{multicols}
\begin{landscape}
\fontsize{8}{8}\selectfont
\begin{longtable}{p{2cm}p{1.75cm}p{3cm}p{1cm}p{1.3cm} p{1.5cm}p{1.5cm}p{1.25cm}p{1.75cm}p{2cm}p{1cm}p{1.5cm}}
\caption{Current easily accessible and freely available multispectral and hyperspectral satellites that are deemed appropriate for precision agriculture \label{tab2}} \\
\hline\hline
		\textbf{Reference}	& \textbf{Name}	& \textbf{Sensor}	& \textbf{Bands} & \textbf{Bandwidth (nm)} & \textbf{Wavelength Range (nm)} & \textbf{Spatial Resolution (m)} & \textbf{Partner Satellite} & \textbf{Repeat Cycle Single Satellite (Days)} & \textbf{Repeat Cycle Multiple Satellite (Days)} & \textbf{Swath Width (km)} & \textbf{Active} \\
\hline\endhead  
\hline\endfoot  
\citet{RN46} & Sentinel-2A	& Multispectral	& 13 & Varies & 420 - 2370 & 10, 20, 60	& Sentinel-2B & 10 & 5 & 290 & 23-06-15 to Present \\
\midrule
\citet{RN46} & Sentinel-2B & Multispectral & 13 & Varies & 420 - 2370 & 10, 20, 60 & Sentinel-2A & 10 & 5 & 290 & 07-03-17 to Present \\
\midrule
\citet{RN47} & Landsat 9 & Multispectral, Panchromatic \& Thermal & 11 & Varies && 30, 15, 100 & Landsat 8 & 16 & 8 & 185 & 27-9-21 to Present \\
\midrule
\citet{RN48} & Landsat 8 & Multispectral, Panchromatic \& Thermal & 11 & Varies && 30, 15, 100 & Landsat 7, Landsat 9 & 16 & 8 & 185 & 11-02-20 to Present \\
\midrule
\citet{RN49} & Landsat 7 & Multispectral, Panchromatic \& Thermal & 8 & Varies & 450 - 2350 & 30, 15, 60 & Landsat 8 & 16 & 8 & 185 & 15-04-99 to 27-09-21 \\
\midrule
\citet{RN50} & KOMPSAT-3 & Multispectral \& Panchromatic & 5 & Varies & 450 - 900 & 2.0 , 0.5 & N/a & 28 & N/A & 16 & 17-05-12 to Present \\
\midrule
\citet{RN50} & KOMPSAT-3A & Multispectral \& Panchromatic & 5 & Varies & 450 - 900 & 1.6 , 0.4 & N/a & 28 & N/A & 13 & 25-03-15 to Present\\
\midrule
\citet{RN51} & Proba-1 (CHRIS) & Hyperspectral & - & 10 & 400 - 1300 & 17 & N/a & 7 & N/A & 14 & 22-10-01 to 4-05-21 \\
\midrule
\citet{RN52} & SPOT-6 & Multispectral \& Panchromatic & 5 & Varies & 450 - 890 & 6, 1.5 & SPOT-7 & 26 & 13 & 60 & 30-06-14 to Present\\
\midrule
\citet{RN52} & SPOT-7 & Multispectral \& Panchromatic & 5 & Varies & 450 - 890 & 6, 1.5 & SPOT-6 & 26 & 13 & 60 & 30-06-14 to Present \\
\midrule
\citet{RN53} & EO1 Hyperion & Hyperspectral & - & 10	& 357 - 2576 & 30 & N/a & 16 & N/A & 7.5  & 21-11-01 to 30-03-17\\
\midrule
\citet{RN54} & PRISMA & Hyperspectral & - & 12 & 400 - 2500 & 30 & N/a & 29 & N/A & 30 & 22-03-19 to Present\\
\midrule
\citet{RN55} & EnMAP & Hyperspectral & - & 6.5, 10 & 420 - 2450 & 30 & N/a & 27 & N/A & 30 & 01-04-22 to Present\\
\end{longtable}
\end{landscape}

\begin{multicols}{2}

\noindent Alternatively, satellites such as Sentinel-2 provide products with atmospheric correction pre-applied or the software to perform it. 	

Vegetation indices have been investigated and adapted as an alternative method to reduce the atmospheric effects. A new vegetation index, Atmospherically Resistant Vegetation Index (ARVI) was proposed to be utilized instead of NDVI with fewer atmospheric effects on the basis that the atmosphere significantly affects the red band \citep{RN45, RN39, RN38}. 

\end{multicols}

\begin{multicols}{2}

\section{Machine Learning Algorithms and Methods of Analysis}
For large-scale sugarcane health monitoring, the common theme amongst current literature is the use of ML algorithms for classification of the plant health into healthy or unhealthy states \citep{RN18, RN19, RN92, RN20, RN95, RN224, RN225, RN226}. Numerous ML methods can be undertaken to solve this classification problem. Table \ref{tab3} provides details of  multispectral and hyperspectral studies in the literature and shows various ML algorithms used in them.

\begin{figure*}[b]
\includegraphics[width=\textwidth]{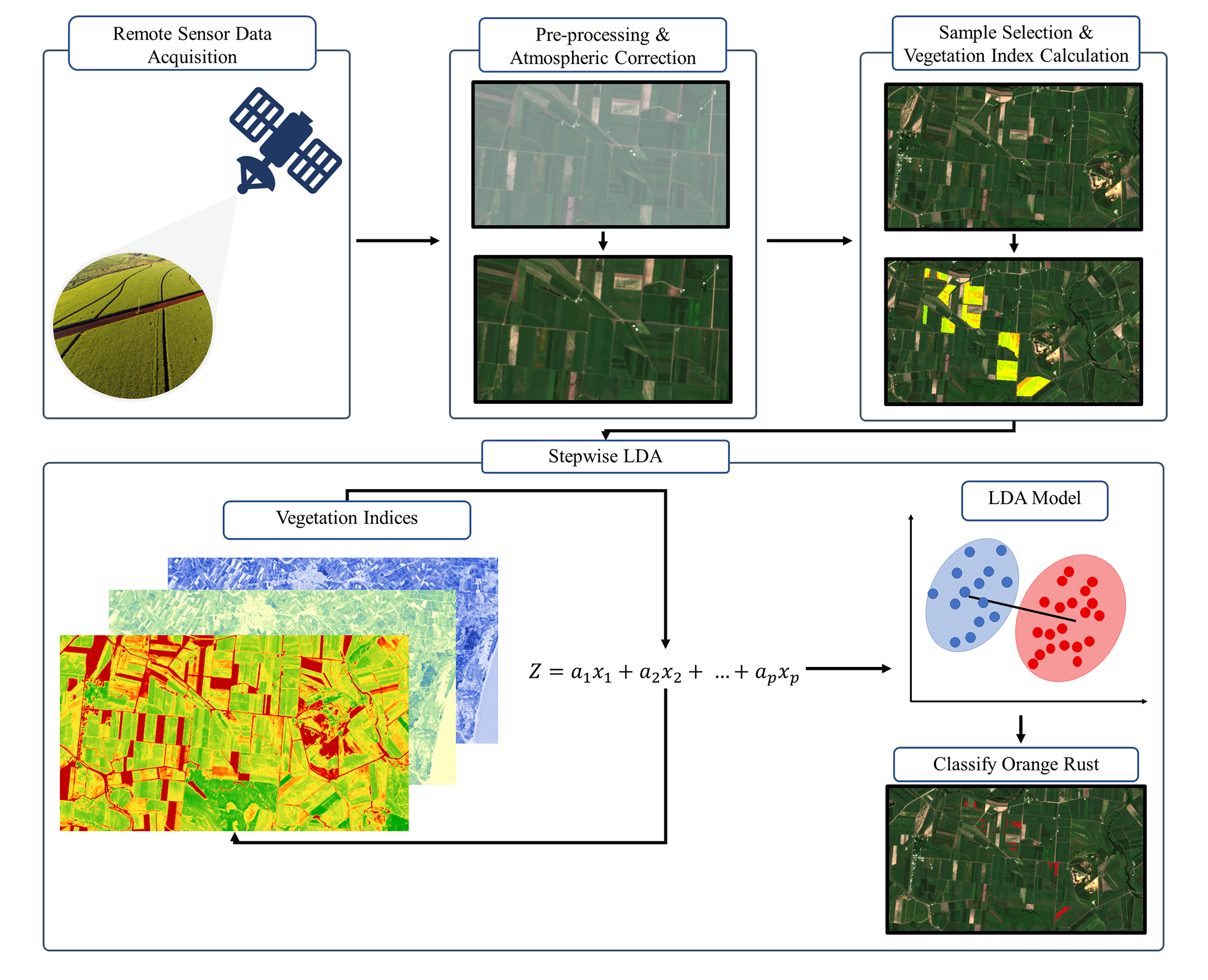}
\caption{Visualization of \citet{RN18, RN19} methodology to classify Orange Rust disease from E0-1 Hyperion images in Mackay, Australia with stepwise LDA. \label{fig12}}
\end{figure*}

The study by \citet{RN18, RN19} classifies orange rust disease in sugarcane exclusively using stepwise Linear Discriminant Analysis (LDA) with vegetation indices as features. This produced a classification accuracy of 96.9\% with the linear combination of DSWI-2, SR695/420, and NDWI-Hyp. This demonstrates that ML algorithms can be effective in classifying the disease state of sugarcane. However, \citet{RN18, RN19} did not compare the effectiveness of different ML algorithms for satellite disease detection, which can be of significance as indicated by other disease detection papers \citep{RN20, RN94, RN224, RN225, RN226}. Additionally, \citet{RN18, RN19} concluded that future studies should be conducted to determine the viability of disease detection in the early stages. A simplified visualisation of the methodology used by \citet{RN18, RN19} can be seen in Figure~\ref{fig12}. The figure demonstrates the overarching approach utilized for spectroscopy health monitoring with ML, in which LDA can be replaced with other ML algorithms.

Several recent studies have made strides in evaluating the effectiveness of various machine learning algorithms for disease detection in sugarcane. Notably, Random Forest and Radial SVM consistently emerged as top performers \citep{RN20, RN224, RN225, RN226}. However, it is important to note that the number of studies comparing machine learning algorithms in this context remains somewhat limited, making it challenging to pinpoint the precise factors contributing to the superior efficacy of a particular method. In a similar vein, \citet{RN33}, while investigating different machine learning algorithms to discern various sugarcane varieties based on their hyperspectral reflectance, found that SVM and Random Forest consistently outperformed other methods. While these findings are still in their early stages, the emergence of Random Forest and Radial SVM as potent tools for health monitoring in sugarcane suggests that non-linear classifiers hold great promise. These classifiers excel at capturing complex patterns and relationships within the data, hinting at the potential for more accurate and robust disease detection systems as this field of research continues to evolve. 

ML algorithms utilized in this context exhibit considerable variability, leading to diverse classification results. For instance, \citet{RN19} employed LDA achieving a classification accuracy of 96.9\%. The effectiveness of LDA hinges on assumptions related to the linearity and separability of spectral data and vegetation indices. However, these assumptions may not consistently hold across different diseases or datasets, and may require statistical analysis to assess whether the data meets the necessary assumptions for LDA implementation \citep{RN84, RN85}. Evaluating the performance of LDA against other models like QDA or SVM with non-linear kernels could shed light on the linear separability of the data. It is very improbable that there is a single machine learning algorithm that will consistently outperform all others for all classification tasks related to sugarcane health monitoring, and consequently the specific characteristics and underlying distributions of the data should be analysed when specifying a desired approach or commenting on its performance. Future studies should transparently present underlying distributions of their data and conduct thorough comparisons of multiple machine learning algorithms based on these characteristics. Furthermore, in alignment with best practices for any machine learning project, rigorous feature selection procedures should be conducted to ascertain the importance of predictors.

\citet{RN92} took a different approach utilising Spectral Information Divergence (SID) to classify mosaic virus in sugarcane from drone-based hyperspectral images achieving a classification accuracy of 92\%. In two instances of implementing Decision Tree (DT) classifiers \citep{RN20, RN94}, sub-optimal performance was exhibited in general or when compared to alternative tree-based methods. Newer alternative tree-based models, such as eXtreme Gradient Boosting (XGB) and Random Forest (RF), should be considered for enhanced accuracy and robustness in sugarcane health monitoring systems. Alternatively, the only other study to develop a DT classifier did not perform cross-validation which is vital to corroborate the effectiveness of the classification method \citep{RN72}. Despite the diverse array of machine learning algorithms explored in existing literature, there has only been one paper on Convolutional Neural Networks (CNNs) in the potential application of neural networks in spectroscopy-based health monitoring for sugarcane \citep{RN479}. Future investigations may benefit from further exploring the capabilities and advantages that neural networks could bring to this domain.

Several studies analysed the underlying distributions of the data and performed a more traditional statistical analysis of the differences between diseased and healthy sugarcane. Canonical Discriminant Analysis (CDA) was implemented by \citet{RN96} and determined that there is a statistically significant difference in the hyperspectral reflectance of Thrips damaged crops compared to healthy crops. A comparable statistical analysis was performed on the spectral reflectance observed with a laboratory-based spectrometer of sugarcane with Brown and Orange Rust by \citet{RN216} demonstrating 11.9\% and 9.9\% differences in the vegetation indices NDVI and ENVI respectively.

\section{Conclusion and Future Work}
In this study, we reviewed factors, identified in the literature, that affect the accuracy of large-scale sugarcane health monitoring, and hence need to be considered in the development of a large-scale sugarcane health monitoring system. This included sugarcane growth factors, vegetation indices, specifications of the satellite-based spectroscopy, factors that influence the observed reflectance of sugarcane, and machine learning algorithms utilized in the analysis. To date, there is limited literature pertaining to a system of this nature and several gaps have been identified for future research. 

A large portion of the current hyperspectral sugarcane health monitoring has been conducted with field spectroscopy with limited use for a large-scale solution for sugarcane farmers. Despite current literature discussing the effectiveness of spectroscopy to detect sugarcane disease/pest, they typically lack a comparison of machine learning algorithms to indicate the most effective classifier for this application. Furthermore, a large portion of known sugarcane health conditions have yet to be detected with newer techniques or systems. There is currently no literature addressing the possibility of classifying diseases and pests at different points in the sugarcane life cycle with satellite-based multi-temporal data.

A characteristic vital to an effective sugarcane health monitoring system is the ability to detect and/or identify multiple diseases, pests, and varieties, simultaneously. Currently, there is limited literature demonstrating this with any form of spectral imaging for sugarcane. The current literature indicates the plausibility of a satellite-based health monitoring system. However, to achieve such a system, several gaps in the literature need to be addressed before the commencement of development. Several recommendations have been made for future studies to fill these gaps.

\begin{itemize}
    \itemsep0em
    \item A practical solution to a large-scale sugarcane health monitoring system will need to be capable of identifying a larger number of health conditions simultaneously. The effectiveness of satellite-based spectroscopy for health monitoring should be evaluated for impactful sugarcane diseases yet to be considered, which include but are not limited to, Sereh, Red Rot, RSD, Smut, Pachymetra, Chlorotic Streak, Yellow Spot, and Fiji Leaf Gall.
    \item A wide variety of different sugarcane varieties are utilized worldwide depending on the climate. An ideal large-scale sugarcane health monitoring system will need to be capable of identifying the aforementioned health conditions for all of the most prominent sugarcane varieties by considering the variation in their spectral reflectance. The effectiveness of satellite-based spectroscopy for this should be evaluated to indicate if region-specific health monitoring systems and models are required based on the local sugarcane varieties.
    \item Future studies should move beyond the prevailing focus on raw spectral measurements or limited use of vegetation indices considering the demonstrated success of studies that incorporate multiple vegetation indices into their models. Success with vegetation indices indicating variation in moisture content indicates that their inclusion in future studies and systems should be contemplated.
    \item Further studies should investigate and perform a direct comparison of disease classification effectiveness with satellite-based spectroscopy and drone-based spectroscopy, to indicate trade-offs in accuracy and real-world logistical consequences associated with each method.
    \item Models for classifying health conditions need to consider the environmental and meteorological variations. Current studies often overlook the influence of annual weather fluctuations, including temperature, humidity, sunlight duration, wind patterns, and precipitation. A comprehensive investigation into how these variations impact the effectiveness of health monitoring methods and understanding the interplay between environmental factors and machine learning models will contribute to the reliability of sugarcane health assessments.
    \item Increasing the chance of early detection of health conditions is vital for reducing the spread or progression of the condition. Therefore, it would be valuable to develop and evaluate the effectiveness of models capable of time series disease detection for prominent sugarcane varieties and diseases. Furthermore, evaluating how the different life cycle stages of sugarcane affect disease detection and the prevalence of exposed soil should be considered.
    \item Given the diverse nature of sugarcane health monitoring tasks and datasets, a one-size-fits-all machine learning algorithm is improbable. Future studies should carefully analyze specific characteristics of their dataset, transparently report distributions, and conduct thorough comparisons among multiple algorithms to identify the most suitable approach. Rigorous feature selection is essential to ensure the relevance and impact of predictors in enhancing model performance.
    \item Neural networks and deep learning have shown unparalleled performance in various applications. Future research can develop new neural networks for sugarcane health monitoring and compare them with the most effective machine learning algorithms for health monitoring and variety classification. 
    \item To incentivise widespread adoption of the discussed precision agriculture monitoring technology, it would be desirable to utilise freely available satellites. However, freely available satellites typically have a worse spectral and spatial resolution in comparison to drones or commercial satellites. To determine how much uncertainty is introduced through the utilisation of satellite-based spectroscopy rather than the superior spatial resolution of drone-based spectroscopy, both methods should be compared and analysed. Furthermore, the minimum spatial and spectral resolutions required to develop a large-scale health monitoring system should be determined so that the running and production costs of this system can be reduced where possible. 
    \item Developing software to automatically acquire satellite images, calculate vegetation indices, and detect disease in sugarcane, displaying the results to a user on a simple user interface and dashboard is another valuable research and development direction. The development of a program that is easy to use would be vital for increasing the probability of its adoption in the industry. 
    \item Performing a cost analysis of the sugarcane industry's current health monitoring practices against the proposed large-scale health monitoring system with freely or commercially available satellites would also be essential for adoption and practice changes.
    
\end{itemize}

\printcredits

\section*{Declaration of AI and AI-assisted technologies in the writing process}
During the preparation of this work the author(s) used  DALL·E 2 in order to contribute towards visualisation. After using this tool/service, the author(s) reviewed and edited the content as needed and take(s) full responsibility for the content of the publication.

\bibliographystyle{cas-model2-names}

\bibliography{cas-refs}

\begin{thebibliography}{135}
\expandafter\ifx\csname natexlab\endcsname\relax\def\natexlab#1{#1}\fi
\providecommand{\url}[1]{\texttt{#1}}
\providecommand{\href}[2]{#2}
\providecommand{\path}[1]{#1}
\providecommand{\DOIprefix}{ }
\providecommand{\ArXivprefix}{arXiv:}
\providecommand{\URLprefix}{URL: }
\providecommand{\Pubmedprefix}{pmid:}
\providecommand{\doi}[1]{\href{http://dx.doi.org/#1}{\path{#1}}}
\providecommand{\Pubmed}[1]{\href{pmid:#1}{\path{#1}}}
\providecommand{\bibinfo}[2]{#2}
\ifx\xfnm\relax \def\xfnm[#1]{\unskip,\space#1}\fi
\bibitem[{Abdel-Rahman et~al.(2010)Abdel-Rahman, Ahmed, van~den Berg and
  Way}]{RN96}
\bibinfo{author}{Abdel-Rahman, E.}, \bibinfo{author}{Ahmed, F.},
  \bibinfo{author}{van~den Berg, M.}, \bibinfo{author}{Way, M.},
  \bibinfo{year}{2010}.
\newblock \bibinfo{title}{Potential of spectroscopic data sets for sugarcane
  thrips (fulmekiola serrata kobus) damage detection}.
\newblock \bibinfo{journal}{International Journal of Remote Sensing}
  \bibinfo{volume}{31}, \bibinfo{pages}{4199--4216}.
\newblock \DOIprefix\doi{10.1080/01431160903241981}.
\bibitem[{Agapiou et~al.(2011)Agapiou, Hadjimitsis, Papoutsa, Alexakis and
  Papadavid}]{RN38}
\bibinfo{author}{Agapiou, A.}, \bibinfo{author}{Hadjimitsis, D.G.},
  \bibinfo{author}{Papoutsa, C.}, \bibinfo{author}{Alexakis, D.D.},
  \bibinfo{author}{Papadavid, G.}, \bibinfo{year}{2011}.
\newblock \bibinfo{title}{The importance of accounting for atmospheric effects
  in the application of ndvi and interpretation of satellite imagery supporting
  archaeological research: The case studies of palaepaphos and nea paphos sites
  in cyprus}.
\newblock \bibinfo{journal}{Remote Sensing} \bibinfo{volume}{3},
  \bibinfo{pages}{2605--2629}.
\newblock \DOIprefix\doi{https://doi.org/10.3390/rs3122605}.
\bibitem[{Agnihotri(1983)}]{RN221}
\bibinfo{author}{Agnihotri, V.P.}, \bibinfo{year}{1983}.
\newblock \bibinfo{title}{Diseases of sugarcane.}
\newblock \bibinfo{journal}{Diseases of sugarcane.} .
\bibitem[{Alves~Varella et~al.(2015)Alves~Varella, Gleriani and dos
  Santos}]{RN35}
\bibinfo{author}{Alves~Varella, C.A.}, \bibinfo{author}{Gleriani, J.M.},
  \bibinfo{author}{dos Santos, R.M.}, \bibinfo{year}{2015}.
\newblock \bibinfo{title}{Precision Agriculture and Remote Sensing}.
  \bibinfo{publisher}{Academic Press}, \bibinfo{address}{San Diego}.
  \bibinfo{type}{book section}~\bibinfo{chapter}{9}.
\newblock pp. \bibinfo{pages}{185--203}.
\newblock \DOIprefix\doi{https://doi.org/10.1016/B978-0-12-802239-9.00009-8}.
\bibitem[{Apan et~al.(2003)Apan, Held, Phinn and Markley}]{RN19}
\bibinfo{author}{Apan, A.}, \bibinfo{author}{Held, A.}, \bibinfo{author}{Phinn,
  S.}, \bibinfo{author}{Markley, J.}, \bibinfo{year}{2003}.
\newblock \bibinfo{title}{Formulation and assessment of narrow-band vegetation
  indices from eo1 hyperion imagery for discriminating sugarcane disease}.
\newblock \bibinfo{journal}{Proceedings of the Spatial Sciences Conference} .
\bibitem[{Apan et~al.(2004)Apan, Held, Phinn and Markley}]{RN18}
\bibinfo{author}{Apan, A.}, \bibinfo{author}{Held, A.}, \bibinfo{author}{Phinn,
  S.}, \bibinfo{author}{Markley, J.}, \bibinfo{year}{2004}.
\newblock \bibinfo{title}{Detecting sugarcane ‘orange rust’ disease using
  eo-1 hyperion hyperspectral imagery}.
\newblock \bibinfo{journal}{International Journal of Remote Sensing}
  \bibinfo{volume}{25}, \bibinfo{pages}{489--498}.
\newblock \DOIprefix\doi{10.1080/01431160310001618031}.
\bibitem[{Bailey and Bechet(1986)}]{RN11}
\bibinfo{author}{Bailey, R.}, \bibinfo{author}{Bechet, G.},
  \bibinfo{year}{1986}.
\newblock \bibinfo{title}{Effect of ratoon stunting disease on the yield and
  components of yield of sugarcane under rainfed conditions}.
\newblock \bibinfo{journal}{Proceedings of the South African Sugar
  Technologists Association} \bibinfo{volume}{60}, \bibinfo{pages}{204--210}.
\bibitem[{Bao et~al.(2021)Bao, Zhou, Bhuiyan, Zia, Ford and Gao}]{RN479}
\bibinfo{author}{Bao, D.}, \bibinfo{author}{Zhou, J.},
  \bibinfo{author}{Bhuiyan, S.A.}, \bibinfo{author}{Zia, A.},
  \bibinfo{author}{Ford, R.}, \bibinfo{author}{Gao, Y.}, \bibinfo{year}{2021}.
\newblock \bibinfo{title}{Early detection of sugarcane smut disease in
  hyperspectral images}.
\newblock \bibinfo{journal}{2021 36th International Conference on Image and
  Vision Computing New Zealand (IVCNZ)} ,
  \bibinfo{pages}{1--6}\DOIprefix\doi{10.1109/IVCNZ54163.2021.9653386}.
\bibitem[{Berding and Hurney(2005)}]{RN30}
\bibinfo{author}{Berding, N.}, \bibinfo{author}{Hurney, A.P.},
  \bibinfo{year}{2005}.
\newblock \bibinfo{title}{Flowering and lodging, physiological-based traits
  affecting cane and sugar yield: What do we know of their control mechanisms
  and how do we manage them?}
\newblock \bibinfo{journal}{Field Crops Research} \bibinfo{volume}{92},
  \bibinfo{pages}{261--275}.
\newblock \DOIprefix\doi{https://doi.org/10.1016/j.fcr.2005.01.015}.
\bibitem[{Blackburn(1998)}]{RN220}
\bibinfo{author}{Blackburn, G.A.}, \bibinfo{year}{1998}.
\newblock \bibinfo{title}{Spectral indices for estimating photosynthetic
  pigment concentrations: A test using senescent tree leaves}.
\newblock \bibinfo{journal}{International Journal of Remote Sensing}
  \bibinfo{volume}{19}, \bibinfo{pages}{657--675}.
\newblock \DOIprefix\doi{https://doi.org/10.1080/014311698215919}.
\bibitem[{Broge and Leblanc(2001)}]{RN461}
\bibinfo{author}{Broge, N.}, \bibinfo{author}{Leblanc, E.},
  \bibinfo{year}{2001}.
\newblock \bibinfo{title}{Comparing prediction power and stability of broadband
  and hyperspectral vegetation indices for estimation of green leaf area index
  and canopy chlorophyll density}.
\newblock \bibinfo{journal}{Remote Sensing of Environment}
  \bibinfo{volume}{76}, \bibinfo{pages}{156--172}.
\newblock \DOIprefix\doi{https://doi.org/10.1016/S0034-4257(00)00197-8}.
\bibitem[{Bégué et~al.(2008)Bégué, Todoroff and Pater}]{RN214}
\bibinfo{author}{Bégué, A.}, \bibinfo{author}{Todoroff, P.},
  \bibinfo{author}{Pater, J.}, \bibinfo{year}{2008}.
\newblock \bibinfo{title}{Multi-time scale analysis of sugarcane within-field
  variability: improved crop diagnosis using satellite time series?}
\newblock \bibinfo{journal}{Precision Agriculture} \bibinfo{volume}{9},
  \bibinfo{pages}{161--171}.
\newblock \DOIprefix\doi{10.1007/s11119-008-9063-3}.
\bibitem[{Canegrowers(2019)}]{RN14}
\bibinfo{author}{Canegrowers}, \bibinfo{year}{2019}.
\newblock \bibinfo{title}{Annual Report}.
\newblock \bibinfo{type}{Report}. Canegrowers.
\bibitem[{Carter and Miller(1994)}]{RN455}
\bibinfo{author}{Carter, G.A.}, \bibinfo{author}{Miller, R.L.},
  \bibinfo{year}{1994}.
\newblock \bibinfo{title}{Early detection of plant stress by digital imaging
  within narrow stress-sensitive wavebands}.
\newblock \bibinfo{journal}{Remote Sensing of Environment}
  \bibinfo{volume}{50}, \bibinfo{pages}{295--302}.
\newblock \URLprefix
  \url{https://www.sciencedirect.com/science/article/pii/0034425794900795},
  \DOIprefix\doi{https://doi.org/10.1016/0034-4257(94)90079-5}.
\bibitem[{Carvalho et~al.(2016)Carvalho, {da Silva}, Munhoz,
  Monteiro-Vitorello, Azevedo, Melotto and Camargo}]{RN477}
\bibinfo{author}{Carvalho, G.}, \bibinfo{author}{{da Silva}, T.},
  \bibinfo{author}{Munhoz, A.}, \bibinfo{author}{Monteiro-Vitorello, C.},
  \bibinfo{author}{Azevedo, R.}, \bibinfo{author}{Melotto, M.},
  \bibinfo{author}{Camargo, L.}, \bibinfo{year}{2016}.
\newblock \bibinfo{title}{Development of a qpcr for leifsonia xyli subsp. xyli
  and quantification of the effects of heat treatment of sugarcane cuttings on
  lxx}.
\newblock \bibinfo{journal}{Crop Protection} \bibinfo{volume}{80},
  \bibinfo{pages}{51--55}.
\newblock \DOIprefix\doi{https://doi.org/10.1016/j.cropro.2015.10.029}.
\bibitem[{Chen et~al.(2020)Chen, Feng, Mo, Mo, Ding and Liu}]{RN200}
\bibinfo{author}{Chen, Y.}, \bibinfo{author}{Feng, L.}, \bibinfo{author}{Mo,
  J.}, \bibinfo{author}{Mo, W.}, \bibinfo{author}{Ding, M.},
  \bibinfo{author}{Liu, Z.}, \bibinfo{year}{2020}.
\newblock \bibinfo{title}{Identification of sugarcane with ndvi time series
  based on hj-1 ccd and modis fusion}.
\newblock \bibinfo{journal}{Journal of the Indian Society of Remote Sensing}
  \bibinfo{volume}{48}, \bibinfo{pages}{249--262}.
\newblock \DOIprefix\doi{10.1007/s12524-019-01042-1}.
\bibitem[{Daughtry et~al.(2000a)Daughtry, Walthall, Kim, {de Colstoun} and
  McMurtrey}]{RN217}
\bibinfo{author}{Daughtry, C.}, \bibinfo{author}{Walthall, C.},
  \bibinfo{author}{Kim, M.}, \bibinfo{author}{{de Colstoun}, E.},
  \bibinfo{author}{McMurtrey, J.}, \bibinfo{year}{2000}a.
\newblock \bibinfo{title}{Estimating corn leaf chlorophyll concentration from
  leaf and canopy reflectance}.
\newblock \bibinfo{journal}{Remote Sensing of Environment}
  \bibinfo{volume}{74}, \bibinfo{pages}{229--239}.
\newblock \DOIprefix\doi{https://doi.org/10.1016/S0034-4257(00)00113-9}.
\bibitem[{Daughtry et~al.(2000b)Daughtry, Walthall, Kim, {de Colstoun} and
  McMurtrey}]{RN452}
\bibinfo{author}{Daughtry, C.}, \bibinfo{author}{Walthall, C.},
  \bibinfo{author}{Kim, M.}, \bibinfo{author}{{de Colstoun}, E.},
  \bibinfo{author}{McMurtrey, J.}, \bibinfo{year}{2000}b.
\newblock \bibinfo{title}{Estimating corn leaf chlorophyll concentration from
  leaf and canopy reflectance}.
\newblock \bibinfo{journal}{Remote Sensing of Environment}
  \bibinfo{volume}{74}, \bibinfo{pages}{229--239}.
\newblock \URLprefix
  \url{https://www.sciencedirect.com/science/article/pii/S0034425700001139},
  \DOIprefix\doi{https://doi.org/10.1016/S0034-4257(00)00113-9}.
\bibitem[{Davis et~al.(1984)Davis, Gillaspie, Vidaver and Harris}]{RN474}
\bibinfo{author}{Davis, M.J.}, \bibinfo{author}{Gillaspie, A.G.},
  \bibinfo{author}{Vidaver, A.K.}, \bibinfo{author}{Harris, R.W.},
  \bibinfo{year}{1984}.
\newblock \bibinfo{title}{Clavibacter: a new genus containing some
  phytopathogenic coryneform bacteria, including clavibacter xyli subsp. xyli
  sp. nov., subsp. nov. and clavibacter xyli subsp. cynodontis subsp. nov.,
  pathogens that cause ratoon stunting disease of sugarcane and bermudagrass
  stunting disease†}.
\newblock \bibinfo{journal}{International Journal of Systematic and
  Evolutionary Microbiology} \bibinfo{volume}{34}, \bibinfo{pages}{107--117}.
\newblock \DOIprefix\doi{https://doi.org/10.1099/00207713-34-2-107}.
\bibitem[{Duft et~al.(2019)Duft, Luciano and Fiorio}]{RN66}
\bibinfo{author}{Duft, D.}, \bibinfo{author}{Luciano, A.},
  \bibinfo{author}{Fiorio, P.}, \bibinfo{year}{2019}.
\newblock \bibinfo{title}{Sentinel-2b and random forest algorithm potential for
  sugarcane varieties identification}.
\newblock \bibinfo{journal}{Proceedings of XX Brazilian Symposium on
  Geoinformatics} , \bibinfo{pages}{188--193}.
\bibitem[{Dutia et~al.(2006)Dutia, Bhatiacharya, Rajak, Chattopadhyay, and and
  Parihar}]{RN72}
\bibinfo{author}{Dutia, S.}, \bibinfo{author}{Bhatiacharya, B.},
  \bibinfo{author}{Rajak, D.}, \bibinfo{author}{Chattopadhyay, C.},
  \bibinfo{author}{and, N.}, \bibinfo{author}{Parihar, J.},
  \bibinfo{year}{2006}.
\newblock \bibinfo{title}{Disease detection in mustard crop using eo-1 hyperion
  satellite data.}
\newblock \bibinfo{journal}{Journal of the Indian Society of Remote Sensing
  (Photonirvachak)} \bibinfo{volume}{34}.
\bibitem[{ElMasry and Sun(2010)}]{RN58}
\bibinfo{author}{ElMasry, G.}, \bibinfo{author}{Sun, D.W.},
  \bibinfo{year}{2010}.
\newblock \bibinfo{title}{Chapter 1 - principles of hyperspectral imaging
  technology}, in: \bibinfo{editor}{Sun, D.W.} (Ed.),
  \bibinfo{booktitle}{Hyperspectral Imaging for Food Quality Analysis and
  Control}. \bibinfo{publisher}{Academic Press}, \bibinfo{address}{San Diego},
  pp. \bibinfo{pages}{3--43}.
\newblock \DOIprefix\doi{https://doi.org/10.1016/B978-0-12-374753-2.10001-2}.
\bibitem[{Emick et~al.(2023)Emick, Babcock, White, Hudak, Domke and
  Finley}]{RN230}
\bibinfo{author}{Emick, E.}, \bibinfo{author}{Babcock, C.},
  \bibinfo{author}{White, G.W.}, \bibinfo{author}{Hudak, A.T.},
  \bibinfo{author}{Domke, G.M.}, \bibinfo{author}{Finley, A.O.},
  \bibinfo{year}{2023}.
\newblock \bibinfo{title}{An approach to estimating forest biomass while
  quantifying estimate uncertainty and correcting bias in machine learning
  maps}.
\newblock \bibinfo{journal}{Remote Sensing of Environment}
  \bibinfo{volume}{295}, \bibinfo{pages}{113678}.
\newblock \DOIprefix\doi{https://doi.org/10.1016/j.rse.2023.113678}.
\bibitem[{{European Space Agency}(a)}]{RN53}
\bibinfo{author}{{European Space Agency}}, a.
\newblock \bibinfo{title}{Earth observing 1}.
\newblock
  \bibinfo{howpublished}{\url{https://www.eoportal.org/satellite-missions/eo-1}}.
\newblock \bibinfo{note}{Accessed: 01/05/2022}.
\bibitem[{{European Space Agency}(b)}]{RN51}
\bibinfo{author}{{European Space Agency}}, b.
\newblock \bibinfo{title}{Proba-1}.
\newblock
  \bibinfo{howpublished}{\url{https://earth.esa.int/eogateway/missions/proba-1?text=Proba-1}}.
\newblock \bibinfo{note}{Accessed: 01/05/2022}.
\bibitem[{{European Space Agency}(c)}]{RN99}
\bibinfo{author}{{European Space Agency}}, c.
\newblock \bibinfo{title}{Quickbird-2}.
\newblock
  \bibinfo{howpublished}{\url{https://earth.esa.int/eogateway/missions/quickbird-2}}.
\newblock \bibinfo{note}{Accessed: 01/05/2022}.
\bibitem[{{European Space Agency}(d)}]{RN46}
\bibinfo{author}{{European Space Agency}}, d.
\newblock \bibinfo{title}{Sentinel-2 overview}.
\newblock
  \bibinfo{howpublished}{\url{https://sentinels.copernicus.eu/web/sentinel/missions/sentinel-2/overview}}.
\newblock \bibinfo{note}{Accessed: 01/05/2022}.
\bibitem[{{European Space Agency}(e)}]{RN52}
\bibinfo{author}{{European Space Agency}}, e.
\newblock \bibinfo{title}{Spot 6 and 7 esa archive}.
\newblock
  \bibinfo{howpublished}{\url{https://earth.esa.int/eogateway/catalog/spot-6-and-7-esa-archive?text=spot+6}}.
\newblock \bibinfo{note}{Accessed: 01/05/2022}.
\bibitem[{Everingham et~al.(2007)Everingham, Lowe, Donald, Coomans and
  Markley}]{RN33}
\bibinfo{author}{Everingham, Y.}, \bibinfo{author}{Lowe, K.H.},
  \bibinfo{author}{Donald, D.}, \bibinfo{author}{Coomans, D.},
  \bibinfo{author}{Markley, J.}, \bibinfo{year}{2007}.
\newblock \bibinfo{title}{Advanced satellite imagery to classify sugarcane crop
  characteristics}.
\newblock \bibinfo{journal}{Agronomy for Sustainable Development}
  \bibinfo{volume}{27}.
\newblock \DOIprefix\doi{10.1051/agro:2006034}.
\bibitem[{Fang and Liang(2014)}]{RN71}
\bibinfo{author}{Fang, H.}, \bibinfo{author}{Liang, S.}, \bibinfo{year}{2014}.
\newblock \bibinfo{title}{Leaf area index models}.
\newblock \DOIprefix\doi{https://doi.org/10.1016/B978-0-12-409548-9.09076-X}.
\bibitem[{Fegan et~al.(1998)Fegan, Croft, Teakle, Hayward and Smith}]{RN478}
\bibinfo{author}{Fegan, M.}, \bibinfo{author}{Croft, B.J.},
  \bibinfo{author}{Teakle, D.S.}, \bibinfo{author}{Hayward, A.C.},
  \bibinfo{author}{Smith, G.R.}, \bibinfo{year}{1998}.
\newblock \bibinfo{title}{Sensitive and specific detection of clavibacter xyli
  subsp. xyli, causal agent of ratoon stunting disease of sugarcane, with a
  polymerase chain reaction-based assay}.
\newblock \bibinfo{journal}{Plant Pathology} \bibinfo{volume}{47},
  \bibinfo{pages}{495--504}.
\newblock \DOIprefix\doi{https://doi.org/10.1046/j.1365-3059.1998.00255.x}.
\bibitem[{Feng et~al.(2020)Feng, He, Cheng, Long and Yuan}]{RN98}
\bibinfo{author}{Feng, X.}, \bibinfo{author}{He, L.}, \bibinfo{author}{Cheng,
  Q.}, \bibinfo{author}{Long, X.}, \bibinfo{author}{Yuan, Y.},
  \bibinfo{year}{2020}.
\newblock \bibinfo{title}{Hyperspectral and multispectral remote sensing image
  fusion based on endmember spatial information}.
\newblock \bibinfo{journal}{Remote Sensing} \bibinfo{volume}{12},
  \bibinfo{pages}{1009}.
\newblock \DOIprefix\doi{https://doi.org/10.3390/rs12061009}.
\bibitem[{{Food and Agriculture Organization of the United
  Nations}(2023)}]{RN7}
\bibinfo{author}{{Food and Agriculture Organization of the United Nations}},
  \bibinfo{year}{2023}.
\newblock \bibinfo{title}{Production / crops and livestock products -
  metadata}.
\newblock \URLprefix \url{https://www.fao.org/faostat/en/#data/QCL}.
\bibitem[{Franke and Menz(2007)}]{RN94}
\bibinfo{author}{Franke, J.}, \bibinfo{author}{Menz, G.}, \bibinfo{year}{2007}.
\newblock \bibinfo{title}{Multi-temporal wheat disease detection by
  multi-spectral remote sensing}.
\newblock \bibinfo{journal}{Precision Agriculture} \bibinfo{volume}{8},
  \bibinfo{pages}{161--172}.
\newblock \DOIprefix\doi{10.1007/s11119-007-9036-y}.
\bibitem[{Galeazzi et~al.(2008)Galeazzi, Sacchetti, Cisbani and Babini}]{RN54}
\bibinfo{author}{Galeazzi, C.}, \bibinfo{author}{Sacchetti, A.},
  \bibinfo{author}{Cisbani, A.}, \bibinfo{author}{Babini, G.},
  \bibinfo{year}{2008}.
\newblock \bibinfo{title}{The prisma program}, in: \bibinfo{booktitle}{IGARSS
  2008 - 2008 IEEE International Geoscience and Remote Sensing Symposium}, pp.
  \bibinfo{pages}{IV -- 105--IV -- 108}.
\newblock \DOIprefix\doi{10.1109/IGARSS.2008.4779667}.
\bibitem[{Galvão et~al.(2005)Galvão, Formaggio and Tisot}]{RN63}
\bibinfo{author}{Galvão, L.S.}, \bibinfo{author}{Formaggio, A.R.},
  \bibinfo{author}{Tisot, D.A.}, \bibinfo{year}{2005}.
\newblock \bibinfo{title}{Discrimination of sugarcane varieties in southeastern
  brazil with eo-1 hyperion data}.
\newblock \bibinfo{journal}{Remote Sensing of Environment}
  \bibinfo{volume}{94}, \bibinfo{pages}{523--534}.
\newblock \DOIprefix\doi{https://doi.org/10.1016/j.rse.2004.11.012}.
\bibitem[{Galvão et~al.(2006)Galvão, Formaggio and Tisot}]{RN64}
\bibinfo{author}{Galvão, L.S.}, \bibinfo{author}{Formaggio, A.R.},
  \bibinfo{author}{Tisot, D.A.}, \bibinfo{year}{2006}.
\newblock \bibinfo{title}{The influence of spectral resolution on
  discriminating brazilian sugarcane varieties}.
\newblock \bibinfo{journal}{International Journal of Remote Sensing}
  \bibinfo{volume}{27}, \bibinfo{pages}{769--777}.
\newblock \DOIprefix\doi{10.1080/01431160500166011}.
\bibitem[{Gamon et~al.(1992)Gamon, Peñuelas and Field}]{RN219}
\bibinfo{author}{Gamon, J.}, \bibinfo{author}{Peñuelas, J.},
  \bibinfo{author}{Field, C.}, \bibinfo{year}{1992}.
\newblock \bibinfo{title}{A narrow-waveband spectral index that tracks diurnal
  changes in photosynthetic efficiency}.
\newblock \bibinfo{journal}{Remote Sensing of Environment}
  \bibinfo{volume}{41}, \bibinfo{pages}{35--44}.
\newblock \DOIprefix\doi{https://doi.org/10.1016/0034-4257(92)90059-S}.
\bibitem[{Gao(1996)}]{RN79}
\bibinfo{author}{Gao, B.c.}, \bibinfo{year}{1996}.
\newblock \bibinfo{title}{Ndwi—a normalized difference water index for remote
  sensing of vegetation liquid water from space}.
\newblock \bibinfo{journal}{Remote Sensing of Environment}
  \bibinfo{volume}{58}, \bibinfo{pages}{257--266}.
\newblock \DOIprefix\doi{https://doi.org/10.1016/S0034-4257(96)00067-3}.
\bibitem[{Genc et~al.(2008)Genc, Genc, Turhan, Smith and Nation}]{RN75}
\bibinfo{author}{Genc, H.}, \bibinfo{author}{Genc, L.},
  \bibinfo{author}{Turhan, H.}, \bibinfo{author}{Smith, S.},
  \bibinfo{author}{Nation, J.}, \bibinfo{year}{2008}.
\newblock \bibinfo{title}{Vegetation indices as indicators of damage by the
  sunn pest (hemiptera: Scutelleridae) to field grown wheat}.
\newblock \bibinfo{journal}{African Journal of Biotechnology}
  \bibinfo{volume}{7}.
\bibitem[{Gers(2014)}]{RN67}
\bibinfo{author}{Gers, C.}, \bibinfo{year}{2014}.
\newblock \bibinfo{title}{Relating remotely sensed multi-temporal landsat 7
  etm+ imagery to sugarcane characteristics}, in: \bibinfo{booktitle}{Proc S
  Afr Sug Technol Ass}, \bibinfo{publisher}{Citeseer}. p.~\bibinfo{pages}{7}.
\bibitem[{Gers(2003)}]{RN68}
\bibinfo{author}{Gers, C.J.}, \bibinfo{year}{2003}.
\newblock \bibinfo{title}{Remotely sensed sugarcane phenological
  characteristics at umfolozi south africa}, in: \bibinfo{booktitle}{IGARSS
  2003. 2003 IEEE International Geoscience and Remote Sensing Symposium.
  Proceedings (IEEE Cat. No.03CH37477)}, pp. \bibinfo{pages}{1010--1012 vol.2}.
\newblock \DOIprefix\doi{10.1109/IGARSS.2003.1293995}.
\bibitem[{Ghai et~al.(2014)Ghai, Singh, Martin, McFarlane, van Antwerpen and
  Rutherford}]{RN476}
\bibinfo{author}{Ghai, M.}, \bibinfo{author}{Singh, V.},
  \bibinfo{author}{Martin, L.}, \bibinfo{author}{McFarlane, S.},
  \bibinfo{author}{van Antwerpen, T.}, \bibinfo{author}{Rutherford, R.},
  \bibinfo{year}{2014}.
\newblock \bibinfo{title}{{A rapid and visual loop‐mediated isothermal
  amplification assay to detect Leifsonia xyli subsp. xyli targeting a
  transposase gene}}.
\newblock \bibinfo{journal}{Letters in Applied Microbiology}
  \bibinfo{volume}{59}, \bibinfo{pages}{648--657}.
\newblock \DOIprefix\doi{10.1111/lam.12327}.
\bibitem[{Gitelson and Merzlyak(1994)}]{RN450}
\bibinfo{author}{Gitelson, A.}, \bibinfo{author}{Merzlyak, M.N.},
  \bibinfo{year}{1994}.
\newblock \bibinfo{title}{Quantitative estimation of chlorophyll-a using
  reflectance spectra: Experiments with autumn chestnut and maple leaves}.
\newblock \bibinfo{journal}{Journal of Photochemistry and Photobiology B:
  Biology} \bibinfo{volume}{22}, \bibinfo{pages}{247--252}.
\newblock \URLprefix
  \url{https://www.sciencedirect.com/science/article/pii/1011134493069634},
  \DOIprefix\doi{https://doi.org/10.1016/1011-1344(93)06963-4}.
\bibitem[{Gitelson et~al.(2003)Gitelson, {Gritz †} and Merzlyak}]{RN457}
\bibinfo{author}{Gitelson, A.A.}, \bibinfo{author}{{Gritz †}, Y.},
  \bibinfo{author}{Merzlyak, M.N.}, \bibinfo{year}{2003}.
\newblock \bibinfo{title}{Relationships between leaf chlorophyll content and
  spectral reflectance and algorithms for non-destructive chlorophyll
  assessment in higher plant leaves}.
\newblock \bibinfo{journal}{Journal of Plant Physiology} \bibinfo{volume}{160},
  \bibinfo{pages}{271--282}.
\newblock \URLprefix
  \url{https://www.sciencedirect.com/science/article/pii/S0176161704704034},
  \DOIprefix\doi{https://doi.org/10.1078/0176-1617-00887}.
\bibitem[{Gitelson et~al.(1996)Gitelson, Kaufman and Merzlyak}]{RN456}
\bibinfo{author}{Gitelson, A.A.}, \bibinfo{author}{Kaufman, Y.J.},
  \bibinfo{author}{Merzlyak, M.N.}, \bibinfo{year}{1996}.
\newblock \bibinfo{title}{Use of a green channel in remote sensing of global
  vegetation from eos-modis}.
\newblock \bibinfo{journal}{Remote Sensing of Environment}
  \bibinfo{volume}{58}, \bibinfo{pages}{289--298}.
\newblock \URLprefix
  \url{https://www.sciencedirect.com/science/article/pii/S0034425796000727},
  \DOIprefix\doi{https://doi.org/10.1016/S0034-4257(96)00072-7}.
\bibitem[{Gitelson et~al.(2002)Gitelson, Kaufman, Stark and Rundquist}]{RN82}
\bibinfo{author}{Gitelson, A.A.}, \bibinfo{author}{Kaufman, Y.J.},
  \bibinfo{author}{Stark, R.}, \bibinfo{author}{Rundquist, D.},
  \bibinfo{year}{2002}.
\newblock \bibinfo{title}{Novel algorithms for remote estimation of vegetation
  fraction}.
\newblock \bibinfo{journal}{Remote sensing of Environment}
  \bibinfo{volume}{80}, \bibinfo{pages}{76--87}.
\bibitem[{Godfray et~al.(2010)Godfray, Beddington, Crute, Haddad, Lawrence,
  Muir, Pretty, Robinson, Thomas and Toulmin}]{RN5}
\bibinfo{author}{Godfray, H.C.J.}, \bibinfo{author}{Beddington, J.R.},
  \bibinfo{author}{Crute, I.R.}, \bibinfo{author}{Haddad, L.},
  \bibinfo{author}{Lawrence, D.}, \bibinfo{author}{Muir, J.F.},
  \bibinfo{author}{Pretty, J.}, \bibinfo{author}{Robinson, S.},
  \bibinfo{author}{Thomas, S.M.}, \bibinfo{author}{Toulmin, C.},
  \bibinfo{year}{2010}.
\newblock \bibinfo{title}{Food security: The challenge of feeding 9 billion
  people}.
\newblock \bibinfo{journal}{Science} \bibinfo{volume}{327},
  \bibinfo{pages}{812--818}.
\newblock \DOIprefix\doi{doi:10.1126/science.1185383}.
\bibitem[{Goel and Qin(1994)}]{RN81}
\bibinfo{author}{Goel, N.S.}, \bibinfo{author}{Qin, W.}, \bibinfo{year}{1994}.
\newblock \bibinfo{title}{Influences of canopy architecture on relationships
  between various vegetation indices and lai and fpar: A computer simulation}.
\newblock \bibinfo{journal}{Remote Sensing Reviews} \bibinfo{volume}{10},
  \bibinfo{pages}{309--347}.
\newblock \DOIprefix\doi{https://doi.org/10.1080/02757259409532252}.
\bibitem[{Goetz et~al.(1985)Goetz, Vane, Solomon and Rock}]{RN56}
\bibinfo{author}{Goetz, A.F.H.}, \bibinfo{author}{Vane, G.},
  \bibinfo{author}{Solomon, J.E.}, \bibinfo{author}{Rock, B.N.},
  \bibinfo{year}{1985}.
\newblock \bibinfo{title}{Imaging spectrometry for earth remote sensing}.
\newblock \bibinfo{journal}{Science} \bibinfo{volume}{228},
  \bibinfo{pages}{1147--1153}.
\newblock \DOIprefix\doi{10.1126/science.228.4704.1147}.
\bibitem[{Gregory and George(2011)}]{RN6}
\bibinfo{author}{Gregory, P.J.}, \bibinfo{author}{George, T.S.},
  \bibinfo{year}{2011}.
\newblock \bibinfo{title}{Feeding nine billion: the challenge to sustainable
  crop production}.
\newblock \bibinfo{journal}{Journal of Experimental Botany} .
\bibitem[{Grisham et~al.(2010)Grisham, Johnson and Zimba}]{RN15}
\bibinfo{author}{Grisham, M.P.}, \bibinfo{author}{Johnson, R.M.},
  \bibinfo{author}{Zimba, P.V.}, \bibinfo{year}{2010}.
\newblock \bibinfo{title}{Detecting sugarcane yellow leaf virus infection in
  asymptomatic leaves with hyperspectral remote sensing and associated leaf
  pigment changes}.
\newblock \bibinfo{journal}{J Virol Methods} \bibinfo{volume}{167},
  \bibinfo{pages}{140--5}.
\newblock \DOIprefix\doi{10.1016/j.jviromet.2010.03.024}.
\bibitem[{{Guzmán Q.} et~al.(2023){Guzmán Q.}, Pinto-Ledezma, Frantz,
  Townsend, Juzwik and Cavender-Bares}]{RN227}
\bibinfo{author}{{Guzmán Q.}, J.A.}, \bibinfo{author}{Pinto-Ledezma, J.N.},
  \bibinfo{author}{Frantz, D.}, \bibinfo{author}{Townsend, P.A.},
  \bibinfo{author}{Juzwik, J.}, \bibinfo{author}{Cavender-Bares, J.},
  \bibinfo{year}{2023}.
\newblock \bibinfo{title}{Mapping oak wilt disease from space using land
  surface phenology}.
\newblock \bibinfo{journal}{Remote Sensing of Environment}
  \bibinfo{volume}{298}, \bibinfo{pages}{113794}.
\newblock \DOIprefix\doi{https://doi.org/10.1016/j.rse.2023.113794}.
\bibitem[{Haboudane et~al.(2002a)Haboudane, Miller, Tremblay, Zarco-Tejada and
  Dextraze}]{RN218}
\bibinfo{author}{Haboudane, D.}, \bibinfo{author}{Miller, J.R.},
  \bibinfo{author}{Tremblay, N.}, \bibinfo{author}{Zarco-Tejada, P.J.},
  \bibinfo{author}{Dextraze, L.}, \bibinfo{year}{2002}a.
\newblock \bibinfo{title}{Integrated narrow-band vegetation indices for
  prediction of crop chlorophyll content for application to precision
  agriculture}.
\newblock \bibinfo{journal}{Remote Sensing of Environment}
  \bibinfo{volume}{81}, \bibinfo{pages}{416--426}.
\newblock \DOIprefix\doi{https://doi.org/10.1016/S0034-4257(02)00018-4}.
\bibitem[{Haboudane et~al.(2002b)Haboudane, Miller, Tremblay, Zarco-Tejada and
  Dextraze}]{RN454}
\bibinfo{author}{Haboudane, D.}, \bibinfo{author}{Miller, J.R.},
  \bibinfo{author}{Tremblay, N.}, \bibinfo{author}{Zarco-Tejada, P.J.},
  \bibinfo{author}{Dextraze, L.}, \bibinfo{year}{2002}b.
\newblock \bibinfo{title}{Integrated narrow-band vegetation indices for
  prediction of crop chlorophyll content for application to precision
  agriculture}.
\newblock \bibinfo{journal}{Remote Sensing of Environment}
  \bibinfo{volume}{81}, \bibinfo{pages}{416--426}.
\newblock \URLprefix
  \url{https://www.sciencedirect.com/science/article/pii/S0034425702000184},
  \DOIprefix\doi{https://doi.org/10.1016/S0034-4257(02)00018-4}.
\bibitem[{Hadjimitsis et~al.(2003)Hadjimitsis, Clayton and Retalis}]{RN44}
\bibinfo{author}{Hadjimitsis, D.}, \bibinfo{author}{Clayton, C.},
  \bibinfo{author}{Retalis, A.}, \bibinfo{year}{2003}.
\newblock \bibinfo{title}{On the darkest pixel atmospheric correction
  algorithm: A revised procedure applied over satellite remotely sensed images
  intended for environmental applications}.
\newblock \bibinfo{journal}{Proceedings of SPIE - The International Society for
  Optical Engineering} \bibinfo{volume}{5239}, \bibinfo{pages}{464--471}.
\newblock \DOIprefix\doi{10.1117/12.511520}.
\bibitem[{Hadjimitsis et~al.(2010)Hadjimitsis, Papadavid, Agapiou,
  Themistocleous, Hadjimitsis, Retalis, Michaelides, Chrysoulakis, Toulios and
  Clayton}]{RN39}
\bibinfo{author}{Hadjimitsis, D.G.}, \bibinfo{author}{Papadavid, G.},
  \bibinfo{author}{Agapiou, A.}, \bibinfo{author}{Themistocleous, K.},
  \bibinfo{author}{Hadjimitsis, M.G.}, \bibinfo{author}{Retalis, A.},
  \bibinfo{author}{Michaelides, S.}, \bibinfo{author}{Chrysoulakis, N.},
  \bibinfo{author}{Toulios, L.}, \bibinfo{author}{Clayton, C.R.I.},
  \bibinfo{year}{2010}.
\newblock \bibinfo{title}{Atmospheric correction for satellite remotely sensed
  data intended for agricultural applications: impact on vegetation indices}.
\newblock \bibinfo{journal}{Natural Hazards and Earth System Sciences}
  \bibinfo{volume}{10}, \bibinfo{pages}{89--95}.
\newblock \DOIprefix\doi{10.5194/nhess-10-89-2010}.
\bibitem[{Hapke(1981)}]{RN61}
\bibinfo{author}{Hapke, B.}, \bibinfo{year}{1981}.
\newblock \bibinfo{title}{Bidirectional reflectance spectroscopy: 1. theory}.
\newblock \bibinfo{journal}{Journal of Geophysical Research: Solid Earth}
  \bibinfo{volume}{86}, \bibinfo{pages}{3039--3054}.
\newblock \DOIprefix\doi{https://doi.org/10.1029/JB086iB04p03039}.
\bibitem[{Hapke and Wells(1981)}]{RN62}
\bibinfo{author}{Hapke, B.}, \bibinfo{author}{Wells, E.}, \bibinfo{year}{1981}.
\newblock \bibinfo{title}{Bidirectional reflectance spectroscopy: 2.
  experiments and observations}.
\newblock \bibinfo{journal}{Journal of Geophysical Research: Solid Earth}
  \bibinfo{volume}{86}, \bibinfo{pages}{3055--3060}.
\newblock \DOIprefix\doi{https://doi.org/10.1029/JB086iB04p03055}.
\bibitem[{Hensley(1971)}]{RN202}
\bibinfo{author}{Hensley, S.D.}, \bibinfo{year}{1971}.
\newblock \bibinfo{title}{Management of sugarcane borer populations in
  louisiana, a decade of change}.
\newblock \bibinfo{journal}{Entomophaga} \bibinfo{volume}{16},
  \bibinfo{pages}{133–146}.
\newblock \DOIprefix\doi{10.1007/BF02370696}.
\bibitem[{Huete et~al.(2002)Huete, Didan, Miura, Rodriguez, Gao and
  Ferreira}]{RN459}
\bibinfo{author}{Huete, A.}, \bibinfo{author}{Didan, K.},
  \bibinfo{author}{Miura, T.}, \bibinfo{author}{Rodriguez, E.},
  \bibinfo{author}{Gao, X.}, \bibinfo{author}{Ferreira, L.},
  \bibinfo{year}{2002}.
\newblock \bibinfo{title}{Overview of the radiometric and biophysical
  performance of the modis vegetation indices}.
\newblock \bibinfo{journal}{Remote Sensing of Environment}
  \bibinfo{volume}{83}, \bibinfo{pages}{195--213}.
\newblock \DOIprefix\doi{https://doi.org/10.1016/S0034-4257(02)00096-2}.
  \bibinfo{note}{the Moderate Resolution Imaging Spectroradiometer (MODIS): a
  new generation of Land Surface Monitoring}.
\bibitem[{Hunt and Rock(1989)}]{RN302}
\bibinfo{author}{Hunt, E.}, \bibinfo{author}{Rock, B.N.}, \bibinfo{year}{1989}.
\newblock \bibinfo{title}{Detection of changes in leaf water content using
  near- and middle-infrared reflectances}.
\newblock \bibinfo{journal}{Remote Sensing of Environment}
  \bibinfo{volume}{30}, \bibinfo{pages}{43--54}.
\newblock \DOIprefix\doi{https://doi.org/10.1016/0034-4257(89)90046-1}.
\bibitem[{Ientilucci and Adler-Golden(2019)}]{RN40}
\bibinfo{author}{Ientilucci, E.J.}, \bibinfo{author}{Adler-Golden, S.},
  \bibinfo{year}{2019}.
\newblock \bibinfo{title}{Atmospheric compensation of hyperspectral data: An
  overview and review of in-scene and physics-based approaches}.
\newblock \bibinfo{journal}{IEEE Geoscience and Remote Sensing Magazine}
  \bibinfo{volume}{7}, \bibinfo{pages}{31--50}.
\newblock \DOIprefix\doi{10.1109/MGRS.2019.2904706}.
\bibitem[{James et~al.(2013)James, Witten, Hastie and Tibshirani}]{RN84}
\bibinfo{author}{James, G.}, \bibinfo{author}{Witten, D.},
  \bibinfo{author}{Hastie, T.}, \bibinfo{author}{Tibshirani, R.},
  \bibinfo{year}{2013}.
\newblock \bibinfo{title}{Classification}. \bibinfo{publisher}{Springer New
  York}, \bibinfo{address}{New York, NY}. \bibinfo{type}{book
  section}~\bibinfo{chapter}{4}.
\newblock pp. \bibinfo{pages}{127--173}.
\newblock \DOIprefix\doi{https://doi.org/10.1007/978-1-4614-7138-7_4}.
\bibitem[{Johansen et~al.(2014)Johansen, Robson, Samson, Sallam, Chandler,
  Eaton, Derby and Jennings}]{RN95}
\bibinfo{author}{Johansen, K.}, \bibinfo{author}{Robson, A.},
  \bibinfo{author}{Samson, P.}, \bibinfo{author}{Sallam, N.},
  \bibinfo{author}{Chandler, K.}, \bibinfo{author}{Eaton, A.},
  \bibinfo{author}{Derby, L.}, \bibinfo{author}{Jennings, J.},
  \bibinfo{year}{2014}.
\newblock \bibinfo{title}{Mapping canegrub damage from high spatial resolution
  satellite imagery}, in: \bibinfo{booktitle}{Proceedings of the 36th
  Conference of the Australian Society of Sugar Cane Technologists, ASSCT
  2014}, pp. \bibinfo{pages}{62--70}.
\bibitem[{Johansen et~al.(2018)Johansen, Sallam, Robson, Samson, Chandler,
  Derby, Eaton and Jennings}]{RN213}
\bibinfo{author}{Johansen, K.}, \bibinfo{author}{Sallam, N.},
  \bibinfo{author}{Robson, A.}, \bibinfo{author}{Samson, P.},
  \bibinfo{author}{Chandler, K.}, \bibinfo{author}{Derby, L.},
  \bibinfo{author}{Eaton, A.}, \bibinfo{author}{Jennings, J.},
  \bibinfo{year}{2018}.
\newblock \bibinfo{title}{Using geoeye-1 imagery for multi-temporal
  object-based detection of canegrub damage in sugarcane fields in queensland,
  australia}.
\newblock \bibinfo{journal}{GIScience \& Remote Sensing} \bibinfo{volume}{55},
  \bibinfo{pages}{285--305}.
\newblock \DOIprefix\doi{10.1080/15481603.2017.1417691}.
\bibitem[{Johnson and Wichern(2014)}]{RN85}
\bibinfo{author}{Johnson, R.A.}, \bibinfo{author}{Wichern, D.W.},
  \bibinfo{year}{2014}.
\newblock \bibinfo{title}{Applied multivariate statistical analysis}.
  volume~\bibinfo{volume}{6}.
\newblock \bibinfo{publisher}{Pearson London, UK:}.
\newblock \DOIprefix\doi{https://doi.org/10.1007/978-3-662-45171-7_14}.
\bibitem[{Jordan(1969a)}]{RN458}
\bibinfo{author}{Jordan, C.F.}, \bibinfo{year}{1969}a.
\newblock \bibinfo{title}{Derivation of leaf-area index from quality of light
  on the forest floor}.
\newblock \bibinfo{journal}{Ecology} \bibinfo{volume}{50},
  \bibinfo{pages}{663--666}.
\newblock \DOIprefix\doi{https://doi.org/10.2307/1936256}.
\bibitem[{Jordan(1969b)}]{RN77}
\bibinfo{author}{Jordan, C.F.}, \bibinfo{year}{1969}b.
\newblock \bibinfo{title}{Derivation of leaf‐area index from quality of light
  on the forest floor}.
\newblock \bibinfo{journal}{Ecology} \bibinfo{volume}{50},
  \bibinfo{pages}{663--666}.
\newblock \DOIprefix\doi{https://doi.org/10.2307/1936256}.
\bibitem[{Julien et~al.(1989)Julien, Irvine and Benda}]{RN23}
\bibinfo{author}{Julien, M.H.R.}, \bibinfo{author}{Irvine, J.E.},
  \bibinfo{author}{Benda, G.T.A.}, \bibinfo{year}{1989}.
\newblock \bibinfo{title}{Sugarcane anatomy, morphology and physiology}.
  \bibinfo{publisher}{Elsevier}, \bibinfo{address}{Amsterdam}.
  \bibinfo{type}{book section}~\bibinfo{chapter}{1}.
\newblock pp. \bibinfo{pages}{1--20}.
\newblock \DOIprefix\doi{https://doi.org/10.1016/B978-0-444-42797-7.50005-3}.
\bibitem[{Kaufman and Tanre(1992)}]{RN74}
\bibinfo{author}{Kaufman, Y.J.}, \bibinfo{author}{Tanre, D.},
  \bibinfo{year}{1992}.
\newblock \bibinfo{title}{Atmospherically resistant vegetation index (arvi) for
  eos-modis}.
\newblock \bibinfo{journal}{IEEE transactions on Geoscience and Remote Sensing}
  \bibinfo{volume}{30}, \bibinfo{pages}{261--270}.
\newblock \DOIprefix\doi{10.1109/36.134076}.
\bibitem[{Keith(2002)}]{RN207}
\bibinfo{author}{Keith, J.C.}, \bibinfo{year}{2002}.
\newblock \bibinfo{title}{Final Report - SRDC Project IPB001 Strategies To
  Control Greyback Canegrub in Early Harvested Ratoon Crops}.
\newblock \bibinfo{type}{Report}. BSES.
\bibitem[{Koike and Gillaspie(1989)}]{RN212}
\bibinfo{author}{Koike, H.}, \bibinfo{author}{Gillaspie, A.G.},
  \bibinfo{year}{1989}.
\newblock \bibinfo{title}{Chapter xix - mosaic}, in: \bibinfo{editor}{Ricaud,
  C.}, \bibinfo{editor}{Egan, B.}, \bibinfo{editor}{Gillaspie, A.},
  \bibinfo{editor}{Hughes, C.} (Eds.), \bibinfo{booktitle}{Diseases of
  Sugarcane}. \bibinfo{publisher}{Elsevier}, \bibinfo{address}{Amsterdam}, pp.
  \bibinfo{pages}{301--322}.
\newblock \DOIprefix\doi{https://doi.org/10.1016/B978-0-444-42797-7.50023-5}.
\bibitem[{{L3Harris Software \& Technology Inc}()}]{RN42}
\bibinfo{author}{{L3Harris Software \& Technology Inc}}, .
\newblock \bibinfo{title}{Envi}.
\newblock
  \bibinfo{howpublished}{\url{https://www.l3harrisgeospatial.com/Software-Technology/ENVI}}.
\newblock \bibinfo{note}{Accessed: 25/04/2022}.
\bibitem[{Liang and Wang(2020)}]{RN37}
\bibinfo{editor}{Liang, S.}, \bibinfo{editor}{Wang, J.} (Eds.),
  \bibinfo{year}{2020}.
\newblock \bibinfo{title}{Chapter 4 - Atmospheric correction of optical
  imagery}. \bibinfo{publisher}{Academic Press}.
\newblock pp. \bibinfo{pages}{131--156}.
\newblock \DOIprefix\doi{https://doi.org/10.1016/B978-0-12-815826-5.00004-0}.
\bibitem[{Louhaichi et~al.(2001)Louhaichi, Borman and Johnson}]{RN464}
\bibinfo{author}{Louhaichi, M.}, \bibinfo{author}{Borman, M.M.},
  \bibinfo{author}{Johnson, D.E.}, \bibinfo{year}{2001}.
\newblock \bibinfo{title}{Spatially located platform and aerial photography for
  documentation of grazing impacts on wheat}.
\newblock \bibinfo{journal}{Geocarto International} \bibinfo{volume}{16},
  \bibinfo{pages}{65--70}.
\bibitem[{Macedo et~al.(2015)Macedo, Macedo, {de Campos}, Novaretti and
  Ferraz}]{RN201}
\bibinfo{author}{Macedo, N.}, \bibinfo{author}{Macedo, D.},
  \bibinfo{author}{{de Campos}, M.B.S.}, \bibinfo{author}{Novaretti, W.R.},
  \bibinfo{author}{Ferraz, L.C.C.}, \bibinfo{year}{2015}.
\newblock \bibinfo{title}{Chapter 5 - management of pests and nematodes}, in:
  \bibinfo{editor}{Santos, F.}, \bibinfo{editor}{Borém, A.},
  \bibinfo{editor}{Caldas, C.} (Eds.), \bibinfo{booktitle}{Sugarcane}.
  \bibinfo{publisher}{Academic Press}, \bibinfo{address}{San Diego}, pp.
  \bibinfo{pages}{89--113}.
\newblock \DOIprefix\doi{https://doi.org/10.1016/B978-0-12-802239-9.00005-0}.
\bibitem[{Magarey(2021)}]{RN3}
\bibinfo{author}{Magarey, R.}, \bibinfo{year}{2021}.
\newblock \bibinfo{title}{Ratoon Stunting Disease}.
\newblock \bibinfo{type}{Report}. SRA.
\bibitem[{Magarey et~al.(2021)Magarey, McHardie, Hession, Cripps, Burgess,
  Spannagle, Sutherland, Di~Bella, Milla, Millar, Schembri, Baxter,
  Hetherington, Turner, Jakins, Quinn, Kalkhoran, Gibbs and Ngo}]{RN12}
\bibinfo{author}{Magarey, R.}, \bibinfo{author}{McHardie, R.},
  \bibinfo{author}{Hession, M.}, \bibinfo{author}{Cripps, G.},
  \bibinfo{author}{Burgess, D.}, \bibinfo{author}{Spannagle, B.},
  \bibinfo{author}{Sutherland, P.}, \bibinfo{author}{Di~Bella, L.},
  \bibinfo{author}{Milla, R.}, \bibinfo{author}{Millar, F.},
  \bibinfo{author}{Schembri, A.}, \bibinfo{author}{Baxter, D.},
  \bibinfo{author}{Hetherington, M.}, \bibinfo{author}{Turner, M.},
  \bibinfo{author}{Jakins, A.}, \bibinfo{author}{Quinn, B.},
  \bibinfo{author}{Kalkhoran, S.}, \bibinfo{author}{Gibbs, L.},
  \bibinfo{author}{Ngo, C.}, \bibinfo{year}{2021}.
\newblock \bibinfo{title}{Incidence and economic effects of ratoon stunting
  disease on the queensland sugarcane industry : Assct peer-reviewed paper}.
\newblock \bibinfo{journal}{Proceedings of the Australian Society of Sugar Cane
  Technologists} \bibinfo{volume}{volume 42}, \bibinfo{pages}{520--526}.
\bibitem[{Magarey et~al.(2004)Magarey, Neilsen and Bull}]{RN9}
\bibinfo{author}{Magarey, R.}, \bibinfo{author}{Neilsen, W.},
  \bibinfo{author}{Bull, J.}, \bibinfo{year}{2004}.
\newblock \bibinfo{title}{The effect of orange rust on sugarcane yield in
  breeding selection trials in central queensland: 1999-2001}, in:
  \bibinfo{booktitle}{2004 Conference of the Australian Society of Sugar Cane
  Technologists held at Brisbane, Queensland, Australia}, pp.
  \bibinfo{pages}{1--6}.
\bibitem[{Magarey et~al.(2022)Magarey, Neilsen and Bull}]{RN36}
\bibinfo{author}{Magarey, R.}, \bibinfo{author}{Neilsen, W.},
  \bibinfo{author}{Bull, J.}, \bibinfo{year}{2022}.
\newblock \bibinfo{title}{Diseases of Australian sugarcane field guide}.
\newblock \bibinfo{publisher}{BSES}.
\bibitem[{Matsuoka and Maccheroni(2015)}]{RN210}
\bibinfo{author}{Matsuoka, S.}, \bibinfo{author}{Maccheroni, W.},
  \bibinfo{year}{2015}.
\newblock \bibinfo{title}{Chapter 6 - disease management}, in:
  \bibinfo{editor}{Santos, F.}, \bibinfo{editor}{Borém, A.},
  \bibinfo{editor}{Caldas, C.} (Eds.), \bibinfo{booktitle}{Sugarcane}.
  \bibinfo{publisher}{Academic Press}, \bibinfo{address}{San Diego}, pp.
  \bibinfo{pages}{115--132}.
\newblock \DOIprefix\doi{https://doi.org/10.1016/B978-0-12-802239-9.00006-2}.
\bibitem[{McFeeters(1996)}]{RN78}
\bibinfo{author}{McFeeters, S.K.}, \bibinfo{year}{1996}.
\newblock \bibinfo{title}{The use of the normalized difference water index
  (ndwi) in the delineation of open water features}.
\newblock \bibinfo{journal}{International Journal of Remote Sensing}
  \bibinfo{volume}{17}, \bibinfo{pages}{1425--1432}.
\newblock \DOIprefix\doi{https://doi.org/10.1080/01431169608948714}.
\bibitem[{Merzlyak et~al.(1999a)Merzlyak, Gitelson, Chivkunova and
  Rakitin}]{RN83}
\bibinfo{author}{Merzlyak, M.N.}, \bibinfo{author}{Gitelson, A.A.},
  \bibinfo{author}{Chivkunova, O.B.}, \bibinfo{author}{Rakitin, V.Y.},
  \bibinfo{year}{1999}a.
\newblock \bibinfo{title}{Non-destructive optical detection of pigment changes
  during leaf senescence and fruit ripening}.
\newblock \bibinfo{journal}{Physiologia Plantarum} \bibinfo{volume}{106},
  \bibinfo{pages}{135--141}.
\newblock \DOIprefix\doi{https://doi.org/10.1034/j.1399-3054.1999.106119.x}.
\bibitem[{Merzlyak et~al.(1999b)Merzlyak, Gitelson, Chivkunova and
  Rakitin}]{RN462}
\bibinfo{author}{Merzlyak, M.N.}, \bibinfo{author}{Gitelson, A.A.},
  \bibinfo{author}{Chivkunova, O.B.}, \bibinfo{author}{Rakitin, V.Y.},
  \bibinfo{year}{1999}b.
\newblock \bibinfo{title}{Non-destructive optical detection of pigment changes
  during leaf senescence and fruit ripening}.
\newblock \bibinfo{journal}{Physiologia plantarum} \bibinfo{volume}{106},
  \bibinfo{pages}{135--141}.
\bibitem[{Merzlyak et~al.(1999c)Merzlyak, Gitelson, Chivkunova and
  Rakitin}]{RN76}
\bibinfo{author}{Merzlyak, M.N.}, \bibinfo{author}{Gitelson, A.A.},
  \bibinfo{author}{Chivkunova, O.B.}, \bibinfo{author}{Rakitin, V.Y.},
  \bibinfo{year}{1999}c.
\newblock \bibinfo{title}{Non‐destructive optical detection of pigment
  changes during leaf senescence and fruit ripening}.
\newblock \bibinfo{journal}{Physiologia plantarum} \bibinfo{volume}{106},
  \bibinfo{pages}{135--141}.
\newblock \DOIprefix\doi{https://doi.org/10.1034/j.1399-3054.1999.106119.x}.
\bibitem[{Moriya et~al.(2017)Moriya, Imai, Tommaselli and Miyoshi}]{RN92}
\bibinfo{author}{Moriya, E.A.S.}, \bibinfo{author}{Imai, N.N.},
  \bibinfo{author}{Tommaselli, A.M.G.}, \bibinfo{author}{Miyoshi, G.T.},
  \bibinfo{year}{2017}.
\newblock \bibinfo{title}{Mapping mosaic virus in sugarcane based on
  hyperspectral images}.
\newblock \bibinfo{journal}{IEEE Journal of Selected Topics in Applied Earth
  Observations and Remote Sensing} \bibinfo{volume}{10},
  \bibinfo{pages}{740--748}.
\newblock \DOIprefix\doi{10.1109/JSTARS.2016.2635482}.
\bibitem[{Moriya et~al.(2018)Moriya, Tommaselli and Imai}]{RN69}
\bibinfo{author}{Moriya, E.A.S.}, \bibinfo{author}{Tommaselli, A.M.G.},
  \bibinfo{author}{Imai, N.N.}, \bibinfo{year}{2018}.
\newblock \bibinfo{title}{A study on the effects of viewing angle variation in
  sugarcane radiometric measures}.
\newblock \bibinfo{journal}{Boletim de Ciencias Geodesicas}
  \bibinfo{volume}{24}, \bibinfo{pages}{85--97}.
\newblock \DOIprefix\doi{10.1590/S1982-21702018000100007}.
\bibitem[{Narmilan et~al.(2022)Narmilan, Gonzalez, Salgadoe and Powell}]{RN20}
\bibinfo{author}{Narmilan, A.}, \bibinfo{author}{Gonzalez, F.},
  \bibinfo{author}{Salgadoe, A.S.A.}, \bibinfo{author}{Powell, K.},
  \bibinfo{year}{2022}.
\newblock \bibinfo{title}{Detection of white leaf disease in sugarcane using
  machine learning techniques over uav multispectral images}.
\newblock \bibinfo{journal}{Drones} \bibinfo{volume}{6}, \bibinfo{pages}{230}.
\newblock \DOIprefix\doi{https://doi.org/10.3390/drones6090230}.
\bibitem[{{NASA}(a)}]{RN49}
\bibinfo{author}{{NASA}}, a.
\newblock \bibinfo{title}{{Landsat Missions}, landsat 7}.
\newblock
  \bibinfo{howpublished}{\url{https://www.usgs.gov/landsat-missions/landsat-7}}.
\newblock \bibinfo{note}{Accessed: 01/05/2022}.
\bibitem[{{NASA}(b)}]{RN48}
\bibinfo{author}{{NASA}}, b.
\newblock \bibinfo{title}{Landsat missions, landsat 8}.
\newblock
  \bibinfo{howpublished}{\url{https://www.usgs.gov/landsat-missions/landsat-8}}.
\newblock \bibinfo{note}{Accessed: 01/05/2022}.
\bibitem[{{NASA}(c)}]{RN47}
\bibinfo{author}{{NASA}}, c.
\newblock \bibinfo{title}{Landsat missions, landsat 9}.
\newblock
  \bibinfo{howpublished}{\url{https://www.usgs.gov/landsat-missions/landsat-9}}.
\newblock \bibinfo{note}{Accessed: 01/05/2022}.
\bibitem[{Nikos~Alexandratos(2012)}]{RN1}
\bibinfo{author}{Nikos~Alexandratos, J.B.}, \bibinfo{year}{2012}.
\newblock \bibinfo{title}{World agriculture towards 2030/2050: the 2012
  revision}.
\newblock \bibinfo{journal}{ESA} \bibinfo{volume}{12-03}.
\newblock \DOIprefix\doi{10.22004/ag.econ.288998}.
\bibitem[{Ong et~al.(2023a)Ong, Jian, Li, Zou, Yin and Ma}]{RN225}
\bibinfo{author}{Ong, P.}, \bibinfo{author}{Jian, J.}, \bibinfo{author}{Li,
  X.}, \bibinfo{author}{Zou, C.}, \bibinfo{author}{Yin, J.},
  \bibinfo{author}{Ma, G.}, \bibinfo{year}{2023}a.
\newblock \bibinfo{title}{New approach for sugarcane disease recognition
  through visible and near-infrared spectroscopy and a modified wavelength
  selection method using machine learning models}.
\newblock \bibinfo{journal}{Spectrochimica Acta Part A: Molecular and
  Biomolecular Spectroscopy} \bibinfo{volume}{302}, \bibinfo{pages}{123037}.
\newblock \DOIprefix\doi{https://doi.org/10.1016/j.saa.2023.123037}.
\bibitem[{Ong et~al.(2023b)Ong, Jian, Li, Zou, Yin and Ma}]{RN226}
\bibinfo{author}{Ong, P.}, \bibinfo{author}{Jian, J.}, \bibinfo{author}{Li,
  X.}, \bibinfo{author}{Zou, C.}, \bibinfo{author}{Yin, J.},
  \bibinfo{author}{Ma, G.}, \bibinfo{year}{2023}b.
\newblock \bibinfo{title}{New approach for sugarcane disease recognition
  through visible and near-infrared spectroscopy and a modified wavelength
  selection method using machine learning models}.
\newblock \bibinfo{journal}{Spectrochimica Acta Part A: Molecular and
  Biomolecular Spectroscopy} \bibinfo{volume}{302}, \bibinfo{pages}{123037}.
\newblock \DOIprefix\doi{https://doi.org/10.1016/j.saa.2023.123037}.
\bibitem[{Pandey et~al.(2023)Pandey, {van Nistelrooij}, Maasakkers, Sutar,
  Houweling, Varon, Tol, Gains, Worden and Aben}]{RN229}
\bibinfo{author}{Pandey, S.}, \bibinfo{author}{{van Nistelrooij}, M.},
  \bibinfo{author}{Maasakkers, J.D.}, \bibinfo{author}{Sutar, P.},
  \bibinfo{author}{Houweling, S.}, \bibinfo{author}{Varon, D.J.},
  \bibinfo{author}{Tol, P.}, \bibinfo{author}{Gains, D.},
  \bibinfo{author}{Worden, J.}, \bibinfo{author}{Aben, I.},
  \bibinfo{year}{2023}.
\newblock \bibinfo{title}{Daily detection and quantification of methane leaks
  using sentinel-3: a tiered satellite observation approach with sentinel-2 and
  sentinel-5p}.
\newblock \bibinfo{journal}{Remote Sensing of Environment}
  \bibinfo{volume}{296}, \bibinfo{pages}{113716}.
\newblock \DOIprefix\doi{https://doi.org/10.1016/j.rse.2023.113716}.
\bibitem[{Qi et~al.(1994)Qi, Chehbouni, Huete, Kerr and Sorooshian}]{RN465}
\bibinfo{author}{Qi, J.}, \bibinfo{author}{Chehbouni, A.},
  \bibinfo{author}{Huete, A.}, \bibinfo{author}{Kerr, Y.},
  \bibinfo{author}{Sorooshian, S.}, \bibinfo{year}{1994}.
\newblock \bibinfo{title}{A modified soil adjusted vegetation index}.
\newblock \bibinfo{journal}{Remote Sensing of Environment}
  \bibinfo{volume}{48}, \bibinfo{pages}{119--126}.
\newblock \URLprefix
  \url{https://www.sciencedirect.com/science/article/pii/0034425794901341},
  \DOIprefix\doi{https://doi.org/10.1016/0034-4257(94)90134-1}.
\bibitem[{Ramouthar et~al.(2013)Ramouthar, McFarlane, Berry and
  Rutherford}]{RN10}
\bibinfo{author}{Ramouthar, P.}, \bibinfo{author}{McFarlane, S.},
  \bibinfo{author}{Berry, S.}, \bibinfo{author}{Rutherford, R.},
  \bibinfo{year}{2013}.
\newblock \bibinfo{title}{Yield loss due to sugarcane yellow leaf virus and its
  prevalence in the south african sugar industry}.
\newblock \bibinfo{journal}{South African Sugar Technologists’ Association} ,
  \bibinfo{pages}{244--254}.
\bibitem[{Rassaby et~al.(2003)Rassaby, Girard, Letourmy, Chaume, Irey,
  Lockhart, Kodja and Rott}]{RN8}
\bibinfo{author}{Rassaby, L.}, \bibinfo{author}{Girard, J.C.},
  \bibinfo{author}{Letourmy, P.}, \bibinfo{author}{Chaume, J.},
  \bibinfo{author}{Irey, M.}, \bibinfo{author}{Lockhart, B.},
  \bibinfo{author}{Kodja, H.}, \bibinfo{author}{Rott, P.},
  \bibinfo{year}{2003}.
\newblock \bibinfo{title}{Impact of sugarcane yellow leaf virus on sugarcane
  yield and juice quality in réunion island}.
\newblock \bibinfo{journal}{European Journal of Plant Pathology}
  \bibinfo{volume}{109}, \bibinfo{pages}{459--466}.
\newblock \DOIprefix\doi{10.1023/A:1024211823306}.
\bibitem[{ReSe(02/03/2022)}]{RN43}
\bibinfo{author}{ReSe}, \bibinfo{year}{02/03/2022}.
\newblock \bibinfo{title}{Rese - remote sensing applications}.
\newblock \URLprefix \url{https://www.rese-apps.com/index.html}.
\bibitem[{de~los Reyes et~al.(2020)de~los Reyes, Langheinrich, Schwind,
  Richter, Pflug, Bachmann, Müller, Carmona, Zekoll and Reinartz}]{RN41}
\bibinfo{author}{de~los Reyes, R.}, \bibinfo{author}{Langheinrich, M.},
  \bibinfo{author}{Schwind, P.}, \bibinfo{author}{Richter, R.},
  \bibinfo{author}{Pflug, B.}, \bibinfo{author}{Bachmann, M.},
  \bibinfo{author}{Müller, R.}, \bibinfo{author}{Carmona, E.},
  \bibinfo{author}{Zekoll, V.}, \bibinfo{author}{Reinartz, P.},
  \bibinfo{year}{2020}.
\newblock \bibinfo{title}{Paco: Python-based atmospheric correction}.
\newblock \bibinfo{journal}{Sensors} \bibinfo{volume}{20}.
\newblock \DOIprefix\doi{10.3390/s20051428}.
\bibitem[{Rondeaux et~al.(1996)Rondeaux, Steven and Baret}]{RN453}
\bibinfo{author}{Rondeaux, G.}, \bibinfo{author}{Steven, M.},
  \bibinfo{author}{Baret, F.}, \bibinfo{year}{1996}.
\newblock \bibinfo{title}{Optimization of soil-adjusted vegetation indices}.
\newblock \bibinfo{journal}{Remote Sensing of Environment}
  \bibinfo{volume}{55}, \bibinfo{pages}{95--107}.
\newblock \URLprefix
  \url{https://www.sciencedirect.com/science/article/pii/0034425795001867},
  \DOIprefix\doi{https://doi.org/10.1016/0034-4257(95)00186-7}.
\bibitem[{Rouse~Jr et~al.(1973)Rouse~Jr, Haas, Schell and Deering}]{RN73}
\bibinfo{author}{Rouse~Jr, J.W.}, \bibinfo{author}{Haas, R.H.},
  \bibinfo{author}{Schell, J.}, \bibinfo{author}{Deering, D.},
  \bibinfo{year}{1973}.
\newblock \bibinfo{title}{Monitoring the vernal advancement and retrogradation
  (green wave effect) of natural vegetation}.
\newblock \bibinfo{type}{Report}. Remote Sensing Center Texas A\&M University.
\bibitem[{Roy(1989)}]{RN60}
\bibinfo{author}{Roy, P.S.}, \bibinfo{year}{1989}.
\newblock \bibinfo{title}{Spectral reflectance characteristics of vegetation
  and their use in estimating productive potential}.
\newblock \bibinfo{journal}{Proceedings: Plant Sciences} \bibinfo{volume}{99},
  \bibinfo{pages}{59--81}.
\bibitem[{Roy and Ravan(1996)}]{RN59}
\bibinfo{author}{Roy, P.S.}, \bibinfo{author}{Ravan, S.A.},
  \bibinfo{year}{1996}.
\newblock \bibinfo{title}{Biomass estimation using satellite remote sensing
  data—an investigation on possible approaches for natural forest}.
\newblock \bibinfo{journal}{Journal of Biosciences} \bibinfo{volume}{21},
  \bibinfo{pages}{535--561}.
\newblock \DOIprefix\doi{https://doi.org/10.1007/BF02703218}.
\bibitem[{de~S.~Rossato~Jr et~al.(2013)de~S.~Rossato~Jr, Costa, Madaleno,
  Mutton, Higley and Fernandes}]{RN203}
\bibinfo{author}{de~S.~Rossato~Jr, J.A.}, \bibinfo{author}{Costa, G.H.G.},
  \bibinfo{author}{Madaleno, L.L.}, \bibinfo{author}{Mutton, M.J.R.},
  \bibinfo{author}{Higley, L.G.}, \bibinfo{author}{Fernandes, O.A.},
  \bibinfo{year}{2013}.
\newblock \bibinfo{title}{Characterization and impact of the sugarcane borer on
  sugarcane yield and quality}.
\newblock \bibinfo{journal}{Agronomy Journal} \bibinfo{volume}{105},
  \bibinfo{pages}{643--648}.
\newblock \DOIprefix\doi{https://doi.org/10.2134/agronj2012.0309}.
\bibitem[{Sahoo et~al.(2015)Sahoo, Ray and R}]{RN57}
\bibinfo{author}{Sahoo, R.}, \bibinfo{author}{Ray, S.}, \bibinfo{author}{R,
  M.}, \bibinfo{year}{2015}.
\newblock \bibinfo{title}{Hyperspectral remote sensing of agriculture}.
\newblock \bibinfo{journal}{Current science} \bibinfo{volume}{108},
  \bibinfo{pages}{848--859}.
\bibitem[{Sallam(2011)}]{RN206}
\bibinfo{author}{Sallam, N.}, \bibinfo{year}{2011}.
\newblock \bibinfo{title}{Review of current knowledge on the population
  dynamics of dermolepida albohirtum (waterhouse) (coleoptera: Scarabaeidae)}.
\newblock \bibinfo{journal}{Australian Journal of Entomology}
  \bibinfo{volume}{50}, \bibinfo{pages}{300--308}.
\newblock \DOIprefix\doi{https://doi.org/10.1111/j.1440-6055.2010.00807.x}.
\bibitem[{Sara et~al.(2021)Sara, Mandava, Kumar, Duela and Jude}]{RN97}
\bibinfo{author}{Sara, D.}, \bibinfo{author}{Mandava, A.K.},
  \bibinfo{author}{Kumar, A.}, \bibinfo{author}{Duela, S.},
  \bibinfo{author}{Jude, A.}, \bibinfo{year}{2021}.
\newblock \bibinfo{title}{Hyperspectral and multispectral image fusion
  techniques for high resolution applications: a review}.
\newblock \bibinfo{journal}{Earth Science Informatics} \bibinfo{volume}{14},
  \bibinfo{pages}{1685 -- 1705}.
\newblock \DOIprefix\doi{https://doi.org/10.1007/s12145-021-00621-6}.
\bibitem[{SIIS()}]{RN50}
\bibinfo{author}{SIIS}, .
\newblock \bibinfo{title}{Kompsat series}.
\newblock \bibinfo{howpublished}{\url{http://www.si-imaging.com/products}}.
\newblock \bibinfo{note}{Accessed: 2010-09-30}.
\bibitem[{Sims and Gamon(2002)}]{RN451}
\bibinfo{author}{Sims, D.A.}, \bibinfo{author}{Gamon, J.A.},
  \bibinfo{year}{2002}.
\newblock \bibinfo{title}{Relationships between leaf pigment content and
  spectral reflectance across a wide range of species, leaf structures and
  developmental stages}.
\newblock \bibinfo{journal}{Remote Sensing of Environment}
  \bibinfo{volume}{81}, \bibinfo{pages}{337--354}.
\newblock \URLprefix
  \url{https://www.sciencedirect.com/science/article/pii/S003442570200010X},
  \DOIprefix\doi{https://doi.org/10.1016/S0034-4257(02)00010-X}.
\bibitem[{Simões and {Rios do Amaral}(2023)}]{RN224}
\bibinfo{author}{Simões, I.O.}, \bibinfo{author}{{Rios do Amaral}, L.},
  \bibinfo{year}{2023}.
\newblock \bibinfo{title}{Uav-based multispectral data for sugarcane resistance
  phenotyping of orange and brown rust}.
\newblock \bibinfo{journal}{Smart Agricultural Technology} \bibinfo{volume}{4},
  \bibinfo{pages}{100144}.
\newblock \DOIprefix\doi{https://doi.org/10.1016/j.atech.2022.100144}.
\bibitem[{Singh et~al.(1997)Singh, Rao, Singh and Singh}]{RN211}
\bibinfo{author}{Singh, S.}, \bibinfo{author}{Rao, G.}, \bibinfo{author}{Singh,
  J.}, \bibinfo{author}{Singh, S.}, \bibinfo{year}{1997}.
\newblock \bibinfo{title}{Effect of sugarcane mosaic potyvirus infection on
  metabolic activity, yield and juice quality}.
\newblock \bibinfo{journal}{Sugar Cane (United Kingdom)} .
\bibitem[{Singh et~al.(2003)Singh, Sinha and Kumar}]{RN209}
\bibinfo{author}{Singh, V.}, \bibinfo{author}{Sinha, O.},
  \bibinfo{author}{Kumar, R.}, \bibinfo{year}{2003}.
\newblock \bibinfo{title}{Progressive decline in yield and quality of sugarcane
  due to sugarcane mosaic virus}.
\newblock \bibinfo{journal}{Indian Phytopathology} \bibinfo{volume}{56},
  \bibinfo{pages}{500--502}.
\bibitem[{Slagter et~al.(2023)Slagter, Reiche, Marcos, Mullissa, Lossou,
  Peña-Claros and Herold}]{RN231}
\bibinfo{author}{Slagter, B.}, \bibinfo{author}{Reiche, J.},
  \bibinfo{author}{Marcos, D.}, \bibinfo{author}{Mullissa, A.},
  \bibinfo{author}{Lossou, E.}, \bibinfo{author}{Peña-Claros, M.},
  \bibinfo{author}{Herold, M.}, \bibinfo{year}{2023}.
\newblock \bibinfo{title}{Monitoring direct drivers of small-scale tropical
  forest disturbance in near real-time with sentinel-1 and -2 data}.
\newblock \bibinfo{journal}{Remote Sensing of Environment}
  \bibinfo{volume}{295}, \bibinfo{pages}{113655}.
\newblock \DOIprefix\doi{https://doi.org/10.1016/j.rse.2023.113655}.
\bibitem[{Soca-Mu\~{n}oz et~al.(2020)Soca-Mu\~{n}oz, Rodr\`{i}guez-Machado,
  Aday-D\`{i}az, Hern\`{a}ndez-Santana and Orozco-Morales}]{RN216}
\bibinfo{author}{Soca-Mu\~{n}oz, J.L.}, \bibinfo{author}{Rodr\`{i}guez-Machado,
  E.}, \bibinfo{author}{Aday-D\`{i}az, O.},
  \bibinfo{author}{Hern\`{a}ndez-Santana, L.}, \bibinfo{author}{Orozco-Morales,
  R.}, \bibinfo{year}{2020}.
\newblock \bibinfo{title}{Spectral signature of brown rust and orange rust in
  sugarcane}.
\newblock \bibinfo{journal}{Revista Facultad de Ingenier\~A\-a Universidad de
  Antioquia} , \bibinfo{pages}{9 --
  20}\DOIprefix\doi{10.17533/udea.redin.20191042}.
\bibitem[{Som-ard et~al.(2021)Som-ard, Atzberger, Izquierdo-Verdiguier, Vuolo
  and Immitzer}]{RN16}
\bibinfo{author}{Som-ard, J.}, \bibinfo{author}{Atzberger, C.},
  \bibinfo{author}{Izquierdo-Verdiguier, E.}, \bibinfo{author}{Vuolo, F.},
  \bibinfo{author}{Immitzer, M.}, \bibinfo{year}{2021}.
\newblock \bibinfo{title}{Remote sensing applications in sugarcane cultivation:
  A review}.
\newblock \bibinfo{journal}{Remote Sensing} \bibinfo{volume}{13},
  \bibinfo{pages}{4040}.
\newblock \DOIprefix\doi{https://doi.org/10.3390/rs13204040}.
\bibitem[{SRA(2022)}]{RN34}
\bibinfo{author}{SRA}, \bibinfo{year}{2022}.
\newblock \bibinfo{title}{VARIETY GUIDE 2022/2023 Herbert Region}.
\newblock \bibinfo{type}{Report}. SRA.
\bibitem[{Storch()}]{RN55}
\bibinfo{author}{Storch, T.}, .
\newblock \bibinfo{title}{{Earth Observation Center}, enmap}.
\newblock
  \bibinfo{howpublished}{\url{https://www.dlr.de/eoc/en/desktopdefault.aspx/tabid-5514/20470_read-47899/}}.
\newblock \bibinfo{note}{Accessed: 01/05/2022}.
\bibitem[{Susantoro et~al.(2019)Susantoro, Wikantika, Harto and
  Suwardi}]{RN215}
\bibinfo{author}{Susantoro, T.M.}, \bibinfo{author}{Wikantika, K.},
  \bibinfo{author}{Harto, A.B.}, \bibinfo{author}{Suwardi, D.},
  \bibinfo{year}{2019}.
\newblock \bibinfo{title}{Monitoring sugarcane growth phases based on satellite
  image analysis (a case study in indramayu and its surrounding, west java,
  indonesia)}.
\newblock \bibinfo{journal}{HAYATI Journal of Biosciences}
  \bibinfo{volume}{26}, \bibinfo{pages}{117}.
\newblock \DOIprefix\doi{10.4308/hjb.26.3.117}.
\bibitem[{Tilman et~al.(2011)Tilman, Balzer, Hill and Befort}]{RN4}
\bibinfo{author}{Tilman, D.}, \bibinfo{author}{Balzer, C.},
  \bibinfo{author}{Hill, J.}, \bibinfo{author}{Befort, B.L.},
  \bibinfo{year}{2011}.
\newblock \bibinfo{title}{Global food demand and the sustainable
  intensification of agriculture}, in: \bibinfo{booktitle}{Proceedings of the
  National Academy of Sciences}, pp. \bibinfo{pages}{20260--20264}.
\newblock \DOIprefix\doi{doi:10.1073/pnas.1116437108}.
\bibitem[{Tucker(1979)}]{RN80}
\bibinfo{author}{Tucker, C.J.}, \bibinfo{year}{1979}.
\newblock \bibinfo{title}{Red and photographic infrared linear combinations for
  monitoring vegetation}.
\newblock \bibinfo{journal}{Remote Sensing of Environment} \bibinfo{volume}{8},
  \bibinfo{pages}{127--150}.
\newblock \DOIprefix\doi{https://doi.org/10.1016/0034-4257(79)90013-0}.
\bibitem[{Vargas et~al.(2016)Vargas, Mendoza, Gómez, Rivero and
  Espinosa}]{RN93}
\bibinfo{author}{Vargas, L.A.O.}, \bibinfo{author}{Mendoza, G.G.},
  \bibinfo{author}{Gómez, R.A.}, \bibinfo{author}{Rivero, N.A.},
  \bibinfo{author}{Espinosa, L.Y.}, \bibinfo{year}{2016}.
\newblock \bibinfo{title}{Characterization of diatraea saccharalis in sugarcane
  (saccharum officinarum) with field spectroradiometry}.
\newblock \bibinfo{journal}{International Journal of Environmental \&
  Agriculture Research (IJOEAR)} .
\bibitem[{Vincini and Frazzi(2011)}]{RN460}
\bibinfo{author}{Vincini, M.}, \bibinfo{author}{Frazzi, E.},
  \bibinfo{year}{2011}.
\newblock \bibinfo{title}{Comparing narrow and broad-band vegetation indices to
  estimate leaf chlorophyll content in planophile crop canopies}.
\newblock \bibinfo{journal}{Precision Agriculture} \bibinfo{volume}{12},
  \bibinfo{pages}{334--344}.
\newblock \DOIprefix\doi{10.1007/s11119-010-9204-3}.
\bibitem[{Walthall et~al.(1985)Walthall, Norman, Welles, Campbell and
  Blad}]{RN70}
\bibinfo{author}{Walthall, C.L.}, \bibinfo{author}{Norman, J.M.},
  \bibinfo{author}{Welles, J.M.}, \bibinfo{author}{Campbell, G.},
  \bibinfo{author}{Blad, B.L.}, \bibinfo{year}{1985}.
\newblock \bibinfo{title}{Simple equation to approximate the bidirectional
  reflectance from vegetative canopies and bare soil surfaces}.
\newblock \bibinfo{journal}{Applied Optics} \bibinfo{volume}{24},
  \bibinfo{pages}{383--387}.
\newblock \DOIprefix\doi{10.1364/AO.24.000383}.
\bibitem[{Wang et~al.(2021)Wang, Cang, Qin, Shan, Zhang, Wang, Li and
  Huang}]{RN301}
\bibinfo{author}{Wang, X.Y.}, \bibinfo{author}{Cang, X.Y.},
  \bibinfo{author}{Qin, W.}, \bibinfo{author}{Shan, H.L.},
  \bibinfo{author}{Zhang, R.Y.}, \bibinfo{author}{Wang, C.M.},
  \bibinfo{author}{Li, W.F.}, \bibinfo{author}{Huang, Y.K.},
  \bibinfo{year}{2021}.
\newblock \bibinfo{title}{Evaluation of field resistance to brown stripe
  disease in novel and major cultivated sugarcane varieties in china}.
\newblock \bibinfo{journal}{Journal of Plant Pathology} \bibinfo{volume}{103},
  \bibinfo{pages}{985--989}.
\newblock \DOIprefix\doi{https://doi.org/10.1007/s42161-021-00870-w}.
\bibitem[{Way et~al.(2006)Way, Leslie, Keeping, Govender et~al.}]{RN204}
\bibinfo{author}{Way, M.}, \bibinfo{author}{Leslie, G.},
  \bibinfo{author}{Keeping, M.}, \bibinfo{author}{Govender, A.}, et~al.,
  \bibinfo{year}{2006}.
\newblock \bibinfo{title}{Incidence of fulmekiola serrata (thysanoptera:
  Thripidae) in south african sugarcane}, in: \bibinfo{booktitle}{Proceedings
  of the South African Sugar Technologists' Association}, pp.
  \bibinfo{pages}{199--201}.
\bibitem[{Way et~al.(2010)Way, Rutherford, Sewpersad, Leslie and
  Keeping}]{RN205}
\bibinfo{author}{Way, M.}, \bibinfo{author}{Rutherford, R.},
  \bibinfo{author}{Sewpersad, C.}, \bibinfo{author}{Leslie, G.},
  \bibinfo{author}{Keeping, M.}, \bibinfo{year}{2010}.
\newblock \bibinfo{title}{Impact of sugarcane thrips, fulmekiola serrata
  (kobus) (thysanoptera: Thripidae) on sugarcane yield in field trials}, in:
  \bibinfo{booktitle}{Proceedings of the South African Sugar Technologists'
  Association}, pp. \bibinfo{pages}{244--256}.
\bibitem[{Xu et~al.(2023)Xu, Li, Du, Mao, Zhou, Huang, Fan, Chen, Ni and
  Guo}]{RN228}
\bibinfo{author}{Xu, Y.}, \bibinfo{author}{Li, X.}, \bibinfo{author}{Du, H.},
  \bibinfo{author}{Mao, F.}, \bibinfo{author}{Zhou, G.},
  \bibinfo{author}{Huang, Z.}, \bibinfo{author}{Fan, W.},
  \bibinfo{author}{Chen, Q.}, \bibinfo{author}{Ni, C.}, \bibinfo{author}{Guo,
  K.}, \bibinfo{year}{2023}.
\newblock \bibinfo{title}{Improving extraction phenology accuracy using sif
  coupled with the vegetation index and mapping the spatiotemporal pattern of
  bamboo forest phenology}.
\newblock \bibinfo{journal}{Remote Sensing of Environment}
  \bibinfo{volume}{297}, \bibinfo{pages}{113785}.
\newblock \DOIprefix\doi{https://doi.org/10.1016/j.rse.2023.113785}.
\bibitem[{Xue and Su(2017)}]{RN45}
\bibinfo{author}{Xue, J.}, \bibinfo{author}{Su, B.}, \bibinfo{year}{2017}.
\newblock \bibinfo{title}{Significant remote sensing vegetation indices: A
  review of developments and applications}.
\newblock \bibinfo{journal}{Journal of Sensors} \bibinfo{volume}{2017},
  \bibinfo{pages}{1353691}.
\newblock \DOIprefix\doi{10.1155/2017/1353691}.
\bibitem[{Yamane(2019)}]{RN17}
\bibinfo{author}{Yamane, T.}, \bibinfo{year}{2019}.
\newblock \bibinfo{title}{Sugarcane}.
\bibitem[{Young et~al.(2016)Young, Kawamata, Ensbey, Lambley and Nock}]{RN475}
\bibinfo{author}{Young, A.J.}, \bibinfo{author}{Kawamata, A.},
  \bibinfo{author}{Ensbey, M.A.}, \bibinfo{author}{Lambley, E.},
  \bibinfo{author}{Nock, C.J.}, \bibinfo{year}{2016}.
\newblock \bibinfo{title}{Efficient diagnosis of ratoon stunting disease of
  sugarcane by quantitative pcr on pooled leaf sheath biopsies}.
\newblock \bibinfo{journal}{Plant Disease} \bibinfo{volume}{100},
  \bibinfo{pages}{2492--2498}.
\newblock \DOIprefix\doi{10.1094/PDIS-06-16-0848-RE}. \bibinfo{note}{pMID:
  30686165}.
\bibitem[{Zabel et~al.(2014)Zabel, Putzenlechner and Mauser}]{RN2}
\bibinfo{author}{Zabel, F.}, \bibinfo{author}{Putzenlechner, B.},
  \bibinfo{author}{Mauser, W.}, \bibinfo{year}{2014}.
\newblock \bibinfo{title}{Global agricultural land resources--a high resolution
  suitability evaluation and its perspectives until 2100 under climate change
  conditions}.
\newblock \bibinfo{journal}{PLoS One} \bibinfo{volume}{9},
  \bibinfo{pages}{e107522}.
\newblock \DOIprefix\doi{10.1371/journal.pone.0107522}.
\bibitem[{Zarco-Tejada et~al.(2005)Zarco-Tejada, Berjón, López-Lozano,
  Miller, Martín, Cachorro, González and {de Frutos}}]{RN463}
\bibinfo{author}{Zarco-Tejada, P.}, \bibinfo{author}{Berjón, A.},
  \bibinfo{author}{López-Lozano, R.}, \bibinfo{author}{Miller, J.},
  \bibinfo{author}{Martín, P.}, \bibinfo{author}{Cachorro, V.},
  \bibinfo{author}{González, M.}, \bibinfo{author}{{de Frutos}, A.},
  \bibinfo{year}{2005}.
\newblock \bibinfo{title}{Assessing vineyard condition with hyperspectral
  indices: Leaf and canopy reflectance simulation in a row-structured
  discontinuous canopy}.
\newblock \bibinfo{journal}{Remote Sensing of Environment}
  \bibinfo{volume}{99}, \bibinfo{pages}{271--287}.
\newblock \DOIprefix\doi{https://doi.org/10.1016/j.rse.2005.09.002}.
\bibitem[{Zhu et~al.(2010)Zhu, Lim, Schenck, Arcinas and Komor}]{RN13}
\bibinfo{author}{Zhu, Y.J.}, \bibinfo{author}{Lim, S.T.S.},
  \bibinfo{author}{Schenck, S.}, \bibinfo{author}{Arcinas, A.},
  \bibinfo{author}{Komor, E.}, \bibinfo{year}{2010}.
\newblock \bibinfo{title}{Rt-pcr and quantitative real-time rt-pcr detection of
  sugarcane yellow leaf virus (scylv) in symptomatic and asymptomatic plants of
  hawaiian sugarcane cultivars and the correlation of scylv titre to yield}.
\newblock \bibinfo{journal}{European Journal of Plant Pathology}
  \bibinfo{volume}{127}, \bibinfo{pages}{263--273}.
\newblock \DOIprefix\doi{10.1007/s10658-010-9591-3}.

\end{thebibliography}
\end{multicols}
\end{document}